\documentclass[conference]{IEEEtran}
\IEEEoverridecommandlockouts
\usepackage{cite}
\usepackage{amsmath,amssymb,amsfonts}
\usepackage{algorithmic}
\usepackage{graphicx}
\usepackage{textcomp}
\usepackage{xcolor}
\usepackage{algorithm}
\usepackage{booktabs}
\usepackage{makecell}
\usepackage{array}
\usepackage{multirow}
\usepackage{placeins}
\usepackage{hyperref}
\usepackage{enumitem}
\usepackage{diagbox}
\usepackage{booktabs}
\usepackage{diagbox}
\usepackage{makecell}
\usepackage{amssymb}
\usepackage{tabularx}
\usepackage{array}
\definecolor{bestred}{RGB}{190,45,40}
\definecolor{secondblue}{RGB}{30,95,170}

\newcommand{\best}[1]{\textbf{\textcolor{bestred}{#1}}}
\newcommand{\secondbest}[1]{\underline{\textit{\textcolor{secondblue}{#1}}}}
\def\BibTeX{{\rm B\kern-.05em{\sc i\kern-.025em b}\kern-.08em
    T\kern-.1667em\lower.7ex\hbox{E}\kern-.125emX}}
\begin{document}

\title{WPBench: A Comprehensive Benchmark\\ for Wind Power Forecasting
}

\author{
    \IEEEauthorblockN{
        Yuhan Zhu$^{1}$, 
        Jilin Hu$^{1}$,  
        Xinying Cai$^{1}$, 
        Yingshan Li$^{1}$,
        Li Ma$^{1}$,
        Xiangfei Qiu$^{1}$\\
        Linsen Li$^{2,3}$,
        Kai Zhang$^{1,3}$,
        Yao Fu$^{3}$,
        Weihao Jiang$^{3}$,
        Bin Yang$^{1}$
    } 
    \IEEEauthorblockA{
        $^1$\textit{School of Data Science \& Engineering, East China Normal University, Shanghai, China}  \\
        $^2$\textit{College of Computer Science and Technology, Zhejiang University, Hangzhou, China} \\
        $^3$\textit{Hangzhou Hikvision Digital Technology Co., Ltd., Hangzhou, China} \\
        $^1$\{yhzhu, 10245501404, 10245501444, 10245501419, xfqiu\}@stu.ecnu.edu.cn, 
        $^1$\{jlhu, byang\}@dase.ecnu.edu.cn \\
        $^3$\{lilinsen, zhangkai23, fuyao, jiangweihao5\}@hikvision.com
    }
} 


\maketitle

\begin{abstract}
Accurate, reliable, and deployable wind power forecasting is critical for power system dispatch, renewable energy integration, and electricity market operations. Progress in this field hinges on the ability to empirically and comprehensively benchmark forecasting methods. Yet existing benchmarks fall short of supporting systematic evaluation in four key aspects: 1) limited coverage of wind power scenarios across turbine scale, variable composition, and spatial structure; 2) incomplete coverage of forecasting model families; 3) evaluation metrics misaligned with wind power requirements; and 4) limited structure-aware diagnostics beyond individual temporal patterns. To address these limitations, we propose WPBench, a comprehensive, fair, and extensible benchmark for wind power forecasting. WPBench integrates 26 public datasets organized by turbine scale and variable composition, spanning single-turbine, multi-turbine, univariate, and multivariate settings. Under unified processing, training, and evaluation protocols, it benchmarks 19 representative models covering traditional methods, deep temporal models, spatio-temporal models, and foundation models. Beyond point-wise errors, WPBench assesses forecast-curve fidelity and computational efficiency, and delivers structure-aware diagnostics across temporal, variable-dependency, and spatial-dependency perspectives. Together, these capabilities enable systematic model comparison across diverse wind scenarios and provide a reusable platform for future research.

\end{abstract}

\section{Introduction}
Wind power has become a pivotal renewable energy source in the global transition toward low-carbon power systems, evolving from a supplementary resource into an indispensable component of modern power grids as installed capacity continues to grow~\cite{cuesta2025review,zhu2024estimation,wang2024bi}. However, wind power generation is highly dependent on wind speed, wind direction, and meteorological conditions, exhibiting strong stochasticity, intermittency, and volatility~\cite{jia2022peer,ying2025condition,jain2024reliability}. Such characteristics bring considerable difficulties to grid stability, supply-demand balancing, and system operation, making accurate and reliable wind power forecasting a fundamental requirement for renewable energy integration and power system management~\cite{jain2024reliability,kan2020theoretical,fan2025domain,ying2025condition}.

Although wind power forecasting can be formulated as a time series prediction problem, wind power data differ fundamentally from ordinary single-sequence time series. Unlike traffic flow, electricity load, or meteorological observations, wind power data are jointly shaped by weather regimes, turbine operating states, wind farm layouts, control strategies, and grid dispatch requirements. As a result, forecasting tasks in this domain vary substantially in structure: some datasets contain only a single turbine with one target power channel; others incorporate SCADA variables, NWP covariates, or additional weather and operating-state features; and wind-farm-level datasets typically involve multiple turbines with intricate spatial and spatio-temporal dependencies. These structural differences directly affect the information available to forecasting models and the dependencies they are expected to capture. Moreover, forecasting objectives extend well beyond minimizing point-wise errors, as predictions must support grid integration, operational scheduling, market participation, and real-time decision making. Together, these properties establish wind power forecasting as a dedicated application scenario rather than a simple subcase of general time series forecasting.

\begin{figure}[t]
\centering
\includegraphics[width=\columnwidth]{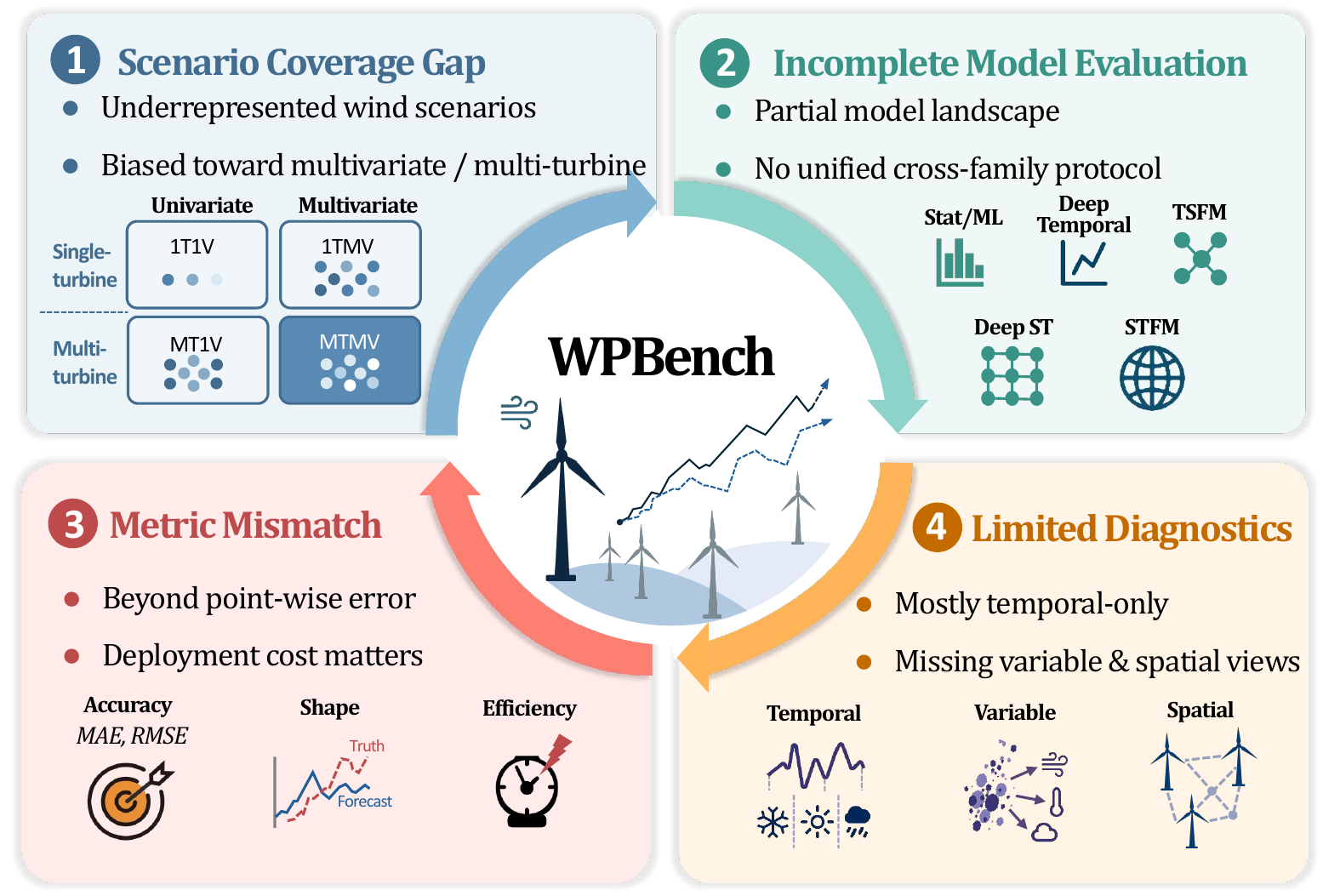}
\caption{In wind power forecasting, existing benchmark frameworks often face four issues: scenario coverage gap, incomplete model-family evaluation, metric mismatch, and limited structure-aware diagnostics.}
\label{fig:wpbench_overview}
\end{figure}

Given this distinctive nature, wind power forecasting cannot be properly evaluated by directly applying general-purpose time series benchmarks. Merely appending a few wind datasets to an existing evaluation suite fails to reflect the diverse data structures, model applicability, and operational constraints that characterize wind power scenarios. This calls for a dedicated and systematic benchmark tailored to wind power forecasting. However, as summarized in Figure~\ref{fig:wpbench_overview}, existing studies and benchmarks remain limited in four key aspects: scenario coverage gap, incomplete model-family evaluation,
metric mismatch, and limited structure-aware diagnostics. We elaborate on each of these limitations in the following.

\noindent\textbf{Issue 1: Limited coverage of wind power scenarios.}
Wind power forecasting is not a single homogeneous scenario. Following the benchmarking insight that different domains exhibit different time-series characteristics\cite{shao2024MTS, qiu2025tab}, wind forecasting also requires coverage of diverse structures within the same domain. Wind datasets can be naturally organized along two axes, the number of turbines and the number of observed variables, yielding four representative scenarios: single-turbine univariate, single-turbine multivariate, multi-turbine univariate, and multi-turbine multivariate forecasting. These scenarios evaluate different model capabilities, including target-series temporal modeling, covariate-aware modeling with SCADA, NWP, or meteorological variables, cross-turbine dependency modeling, and their combination. Thus, wind scenario coverage is not merely about including more wind datasets, but about covering structurally different tasks that reflect different forecasting requirements.

General-purpose TSF benchmarks such as TFB\cite{qiu2024tfb}, fev-bench\cite{shchur2025fev}, and GIFT-Eval\cite{aksu2024gift} include broad cross-domain forecasting tasks and some wind-related datasets or tasks. However, these wind tasks are usually treated as ordinary time series or multivariate time series within a general evaluation suite, rather than being organized around wind-specific structural diversity. In particular, they lack a systematic multi-turbine perspective for evaluating spatial relationships among turbines and the corresponding spatio-temporal or graph-based modeling requirements. Within the wind domain, public resources such as SDWPF\cite{zhou2024sdwpf}, GEFCom 2012\cite{hong2014gefcom2012}, GEFCom 2014\cite{hong2016gefcom2014}, and OpenWPF\cite{xu2025cross} have advanced wind forecasting evaluation, but their coverage remains concentrated on specific settings, such as large-scale multi-turbine wind-farm forecasting, multivariate wind forecasting, or neural cross-dataset evaluation. They still lack balanced structural contrasts across target-only temporal modeling, covariate-aware modeling, spatial dependency modeling, and the combined multi-turbine multivariate setting. This imbalance makes it difficult to separate model weakness from dataset-structure bias, and limits analysis of how different wind scenarios affect forecasting requirements and model behavior.

\noindent\textbf{Issue 2: Incomplete coverage of forecasting model families.}
The diversity of wind data structures also complicates model selection. Single-turbine tasks can often be modeled as ordinary temporal forecasting problems, where statistical models, machine learning methods, and deep temporal models use historical power, SCADA variables, or meteorological covariates\cite{szostek2024analysis,shi2025wind,fan2024dewp}. In contrast, multi-turbine, multi-farm, and regional tasks contain spatial correlations induced by turbine locations, wind propagation, wake effects, and similar operating states, making spatio-temporal or graph-based models relevant\cite{daenens2025spatio,yang2024wind,liang2024wpformer, ma2024lspt}. Existing studies have explored these directions separately, but few benchmarks compare them under the same data processing, training, and evaluation protocol.

As a result, current WPF benchmarks provide only a partial view of model suitability. Statistical WPF benchmarks mainly focus on statistical models\cite{sopena2023benchmarking}; OpenWPF\cite{xu2025cross} promotes cross-dataset evaluation for neural networks, but still emphasizes deep temporal models rather than jointly covering temporal and spatio-temporal paradigms. Meanwhile, time-series foundation models and spatio-temporal foundation models are rapidly developing and claim strong domain generalization ability\cite{kottapalli2025foundation,liang2025foundation, liu2025spatiotemporal, zhang2025timeraf}, yet they have not been systematically evaluated for wind forecasting. A comprehensive wind benchmark should therefore cover statistical learning, machine learning, deep temporal models, deep spatio-temporal models, time-series foundation models, and spatio-temporal foundation models under a unified protocol.

\begin{figure}[t]
\centering
\includegraphics[width=\columnwidth]{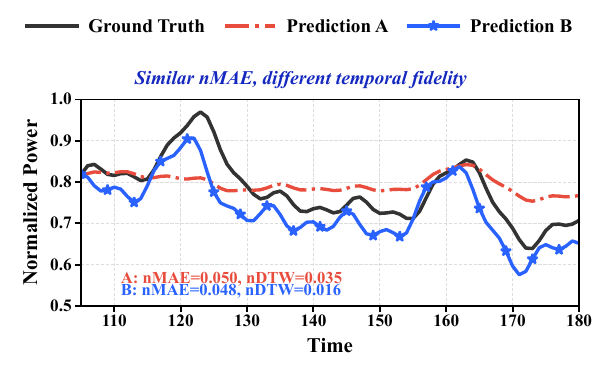}
\caption{Point-wise error alone can be insufficient for wind power forecasting evaluation. Predictions with similar MAE may still differ in fluctuation alignment and forecast-curve fidelity.}
\label{fig:metric_mismatch_intro}
\end{figure}

\noindent\textbf{Issue 3: Metrics misaligned with wind-power requirements.}
Point-wise error metrics such as MAE, RMSE, MAPE, nMAE, and nRMSE are widely used in WPF studies~\cite{piotrowski2022evaluation}. However, low point-wise error does not guarantee that the predicted curve preserves critical wind power patterns. As illustrated in Figure~\ref{fig:metric_mismatch_intro}, two predictions with similar MAE may still differ substantially in the temporal alignment of fluctuations and forecast-curve fidelity. Prior TSF research has similarly shown that similar MSE values may correspond to very different forecast shapes, especially for abrupt or non-stationary sequences, where point-wise metrics can miss differences in fluctuation alignment and curve shape~\cite{Vincent2019Shape}. This issue is more consequential in wind forecasting, where operators care whether forecasts track fluctuations, preserve trend shapes, and capture ramp events that affect dispatch and grid operation~\cite{gallego2015review,meng2025evaluating}.

In addition, WPF is a deployment-oriented industrial task. As deep temporal models, spatio-temporal models, and foundation models are introduced, models differ substantially in training cost, inference latency, memory usage, and parameter size. A model with slightly better numerical accuracy may still be unsuitable for real-time forecasting or large-scale wind-farm deployment if its computational cost is too high. Therefore, wind benchmarks should evaluate models in terms of numerical accuracy, forecast-curve fidelity, and efficiency.


\noindent\textbf{Issue 4: Limited structure-aware diagnostics beyond individual temporal patterns.}
Structure-aware diagnostics aim to explain how data-structural factors in a forecasting task affect model behavior, rather than only reporting average errors or rankings. In wind power forecasting, this means analyzing not only the temporal patterns of individual target power series, but also the relationships between auxiliary variables and power output, as well as spatial relationships among turbines. Recent TSF benchmarks have begun to move toward diagnostic benchmarking. For example, TFB\cite{qiu2024tfb} and TIME\cite{qiao2026time} analyze temporal properties of target series, such as trend, seasonality, stability, and predictability, together with variable correlations. These analyses are valuable, but they still provide a limited view of wind forecasting because they do not explicitly diagnose how covariate relationships and turbine-level spatial structures affect model behavior.

This perspective is not sufficient to describe the structure of wind forecasting tasks. For example, two wind datasets may have target power series with similar seasonality or volatility, but differ in whether they provide informative covariates such as wind speed or NWP variables, or whether they contain meaningful cross-turbine dependencies. Thus, task difficulty is not determined only by non-stationarity, volatility or seasonality of the target series; it also depends on target--covariate relationships and turbine-level spatial dependencies. A wind benchmark should therefore extend dataset characterization beyond individual temporal patterns and organize diagnostics across richer dimensions, including temporal patterns of target power series, variable dependencies between power and auxiliary covariates, and spatial relationships among turbines. Such structure-aware diagnostics can better explain why temporal, spatio-temporal, channel-independent, channel-dependent, and foundation models behave differently under different wind data organizations.

To address these limitations, we present \textbf{WPBench}, a comprehensive and fair benchmark for wind power forecasting. WPBench treats wind power forecasting as a first-class forecasting scenario rather than a small subset of general time series prediction. It organizes open wind datasets according to wind-specific data structures, evaluates broad forecasting model families under a unified protocol, and assesses performance from multiple perspectives that reflect both forecasting accuracy and operational requirements. In addition, WPBench provides structure-aware diagnostics to better understand how temporal patterns, variable dependencies, and spatial structures affect different forecasting models. In this way, WPBench is designed not only to rank methods, but also to clarify which types of models are reliable under different wind data organizations. The benchmark artifact further supports reproducible comparison and future extension with new datasets, models, metrics, and diagnostic protocols. 

Our contributions are summarized as follows.
\begin{itemize}[left=0.3cm]
\item \textbf{Comprehensive Wind Power Dataset Organization.}
We curate and organize 26 open-source wind power datasets along two dimensions—number of turbines and number of variables—yielding a balanced coverage of single-turbine univariate, single-turbine multivariate, multi-turbine univariate, and multi-turbine multivariate settings. This is the first benchmark to explicitly cover all four structural categories, enabling controlled analysis of how data organization affects model behavior.

\item \textbf{Comprehensive Evaluation of Model Families.}
We benchmark 19 representative models across five families—statistical/ML models, deep temporal models, time-series foundation models, deep spatio-temporal models, and spatio-temporal foundation models—under unified experimental protocols. This is the most comprehensive model-family comparison in wind power forecasting to date, revealing which model types are suited to which data structures.

\item \textbf{Multi-dimensional Evaluation Aligned with Operational Requirements.}
We adopt a diverse set of evaluation metrics that go beyond point-wise errors to jointly assess numerical accuracy, forecast-curve fidelity, and computational efficiency. Unlike existing benchmarks that rely solely on MAE/MSE, these metrics reflect forecasting quality dimensions that directly matter for grid dispatch, market participation, and real-time deployment.

\item \textbf{Structure-aware Diagnostic Analysis.}
We provide systematic diagnostic experiments that attribute model performance to three structural factors: temporal patterns in target power series, dependencies between power output and auxiliary covariates, and spatial correlations among turbines. These diagnostics produce interpretable explanations of why certain model families succeed or fail under specific wind data organizations.

\end{itemize}


\section{Related Work}
\subsection{Time Series and Spatio-Temporal Forecasting Benchmarks}

Time series describe temporally ordered numerical measurements and serve as a fundamental data representation for capturing the dynamic behavior of real-world systems~\cite{wu2026timeart}. Such data arise extensively in practical domains, including energy, transportation, environmental monitoring, industrial applications, and healthcare~\cite{wu2025k2vae,qiu2026dag,li2026gcgnet,qiu2026bridging,liu2026rethinking,liu2026astgi,wu2025srsnet,qiu2025dbloss,cheng2026metagnsdformer}. Research on time series covers a variety of analytical problems, ranging from forecasting~\cite{11002729,yu2025merlin} and anomaly detection~\cite{qiu2025tab,wu2025catch,DBLP:journals/corr/abs-2510-18998} to other temporal modeling tasks. Among these problems, multivariate time series forecasting has received substantial attention, with the goal of estimating future values of multiple variables based on their past observations. As forecasting techniques continue to grow in both number and diversity, unified and dependable evaluation frameworks are increasingly needed to support fair model comparison and to reveal the respective strengths and weaknesses of different approaches.

To this end, time series forecasting benchmarks provide a unified basis for evaluating forecasting methods and have played an important role in promoting standardized model comparison. Early benchmarks, such as M3\cite{makridakis2000m3}, M4\cite{makridakis2018m4}, and Monash\cite{godahewa2021monash}, mainly focused on univariate forecasting and cross-domain dataset collection. Later benchmarks, including LTSF-Linear\cite{zeng2023LTSF-Linear}, TSlib\cite{wei2022TSlib}, BasicTS\cite{liang2022basicts}, and TFB\cite{qiu2024tfb}, extended evaluation to long-term forecasting, multivariate forecasting, and deep learning methods. With the rise of time-series foundation models and realistic forecasting evaluation, GIFT-Eval\cite{aksu2024gift}, TSFM-Bench\cite{li2025tsfm}, fev-bench\cite{shchur2025fev}, and TIME\cite{qiao2026time} further emphasize zero-shot or few-shot evaluation\cite{cheng2025gaussian}, cross-domain generalization, covariate-rich tasks, realistic task settings, and diagnostic analysis.

Spatio-temporal forecasting benchmarks evaluate models on data with both temporal dynamics and spatial structures\cite{huang2024benchtemp}. For example, LibCity\cite{wang2021libcity} focuses on urban spatio-temporal prediction, while BasicTS\cite{liang2022basicts} and BasicTS+\cite{shao2024basictsplus} provide unified support for multivariate and spatio-temporal forecasting methods. These benchmarks show the importance of jointly modeling temporal evolution and spatial dependencies. However, their scenarios are still largely concentrated on broad domains such as traffic, urban computing, and weather forecasting. Wind power forecasting therefore remains underrepresented as a distinct energy-domain scenario.

\subsection{Wind Power Forecasting Datasets and Benchmarks}

Several public datasets, competitions, and benchmarks have promoted standardized comparison in wind power forecasting. GEFCom 2012/2014 provided classical wind forecasting competition tasks\cite{hong2014gefcom2012,hong2016gefcom2014}; SDWPF\cite{zhou2024sdwpf} released a large-scale multi-turbine dataset with SCADA variables and turbine spatial information; OpenWPF integrated multiple public wind power datasets for cross-dataset evaluation of neural network-based WPF\cite{xu2025cross}; and statistical WPF benchmarks have supported standardized comparison of statistical forecasting models\cite{sopena2023benchmarking}. These efforts have advanced data openness and comparative evaluation in the wind forecasting community.

Nevertheless, existing wind-domain resources are usually centered on specific task settings, partial data structures, or limited model families. Table~\ref{tab:benchmark_comparison} summarizes the positioning of WPBench against representative forecasting benchmarks.

\begin{table}[t]
\centering
\caption{Comparison of WPBench with representative forecasting benchmarks.}
\label{tab:benchmark_comparison}
\footnotesize
\renewcommand{\arraystretch}{0.92}
\setlength{\tabcolsep}{1.4pt}

\begin{tabularx}{\columnwidth}{@{}
>{\raggedright\arraybackslash}p{0.28\columnwidth}
*{7}{>{\centering\arraybackslash}X}
@{}}
\toprule
Benchmark
& \makecell{Wind-\\spec.}
& \makecell{Struct.\\cov.}
& \makecell{Temp.\\models}
& \makecell{ST\\models}
& \makecell{Found.\\models}
& \makecell{Shape\\eval.}
& \makecell{Struct.\\diag.} \\
\midrule
TFB\cite{qiu2024tfb}
& $\times$ & $\times$ & $\checkmark$ & $\times$ & $\times$ & $\times$ & $\circ$ \\
fev-bench\cite{shchur2025fev}
& $\times$ & $\times$ & $\checkmark$ & $\times$ & $\checkmark$ & $\times$ & $\circ$ \\
GIFT-Eval\cite{aksu2024gift}
& $\times$ & $\times$ & $\checkmark$ & $\times$ & $\checkmark$ & $\times$ & $\circ$ \\
TIME\cite{qiao2026time}
& $\times$ & $\times$ & $\times$ & $\times$ & $\checkmark$ & $\times$ & $\circ$ \\
OpenWPF\cite{xu2025cross}
& $\checkmark$ & $\circ$ & $\checkmark$ & $\times$ & $\times$ & $\times$ & $\times$ \\
SDWPF\cite{zhou2024sdwpf}
& $\checkmark$ & $\circ$ & $\times$ & $\times$ & $\times$ & $\times$ & $\times$ \\
Statistical WPF\cite{sopena2023benchmarking}
& $\checkmark$ & $\circ$ & $\checkmark$ & $\times$ & $\times$ & $\times$ & $\circ$ \\
\textbf{WPBench}
& $\checkmark$ & $\checkmark$ & $\checkmark$ & $\checkmark$ & $\checkmark$ & $\checkmark$ & $\checkmark$ \\
\bottomrule
\multicolumn{8}{@{}p{\columnwidth}@{}}{\footnotesize
$\times$: absent; $\checkmark$: present; $\circ$: partial or limited support.}
\end{tabularx}
\end{table}

\subsection{Wind Power Forecasting Methods}

Wind power forecasting has been studied under diverse modeling paradigms. One line of work formulates WPF as a temporal or multivariate temporal forecasting problem, where historical power records, SCADA variables, NWP covariates, and meteorological observations are used to model temporal dependencies, variable interactions, and nonlinear dynamics~\cite{qiu2026dag,li2026gcgnet}. Early approaches primarily relied on statistical models and classical machine learning methods~\cite{jiang2017statistical}. With the rapid advances of deep learning and its strong representation learning capability demonstrated across areas such as computer vision and video generation~\cite{ma2024followpose,ma2024followyouremoji,ma2025controllable,ma2026livelight}, deep neural networks have been increasingly adopted for wind power forecasting. Representative deep forecasting architectures include LSTM, GRU, CNN, TCN, and Transformer-based models~\cite{yin2019cascaded,xiao2023boostedgru,gong2023cnninformer}.

Another line of work focuses on multi-turbine, multi-farm, or regional forecasting, where turbines, wind farms, or regions are represented as nodes and graph neural networks, spatio-temporal graph models, or spatio-temporal Transformers are used to capture cross-node dependencies\cite{song2022gcnwind,liang2024wpformer,yi2024DCGST,liu2025spatiotemporal, liu2026astgi}. Recent studies such as PowerMistral and WindLLM also explore pre-trained large models for WPF\cite{meng2026powermistral,fan2025windllm}.

These method families are often evaluated in separate experimental settings, making it difficult to compare temporal models, spatio-temporal models, and foundation models under a unified protocol. WPBench is complementary to existing forecasting benchmarks and wind-domain studies: it focuses on organizing open wind datasets by data structure, evaluating broad model families consistently, and supporting structure-aware diagnostics for wind-specific forecasting scenarios.


\section{Preliminaries}
\subsection{Wind Power Forecasting Setting}
Let $\mathcal{X}\in\mathbb{R}^{N\times T\times C}$ denote a wind power time series dataset, where $N$ is the number of turbines or spatially separated wind power series, $T$ is the number of time steps, and $C$ is the number of observed channels. The first channel is the target wind power series, denoted by $\mathbf{Y}\in\mathbb{R}^{N\times T}$, and the remaining channels, when available, are auxiliary covariates such as wind speed, wind direction, temperature, or turbine operating states. Given a look-back window of length $L$, the forecasting task is to predict
the target power values over the next $H$ steps, where
$\mathbf{Y}_{:,t+1:t+H}\in\mathbb{R}^{N\times H}$:
\begin{equation}
    \widehat{\mathbf{Y}}_{:,t+1:t+H}
    =
    f_{\theta}(\mathcal{X}_{:,t-L+1:t,:}),
\end{equation}
where $H$ is the forecasting horizon. This unified formulation covers single-turbine and multi-turbine datasets, as well as settings with or without auxiliary covariates.

\subsection{Dataset Characterization Metrics}
\label{sec:dataset_characterization_metrics}
Existing forecasting benchmarks have shown that dataset characteristics help explain why different models succeed under different temporal patterns\cite{qiu2024tfb,qiao2026time}. For wind power forecasting, however, the relevant structure is not limited to the target power sequence. A dataset may also contain auxiliary SCADA or meteorological variables, multiple turbine nodes, and spatially organized power series. WPBench therefore characterizes each dataset from three perspectives: temporal patterns of the target power series, dependencies between target power and auxiliary variables, and spatial dependencies among turbine targets.

\subsubsection{Temporal Characteristics}
For each turbine, the first column in its variable group is treated as the target power series $x\in\mathbb{R}^{T}$. Following the common practice of time-series characterization, WPBench uses STL decomposition to separate recurring, long-term, and residual components:
\begin{equation}
    x_t = S_t + G_t + R_t,
\end{equation}
where $S_t$, $G_t$, and $R_t$ denote the seasonal, trend, and residual components, respectively. Candidate periods are generated from FFT peaks and a default period set, and the decomposition with the strongest seasonal component is retained. 

\noindent\textbf{Seasonality.}
Seasonality measures the strength of recurring patterns in the target power series and is computed as an explained-variance ratio:
\begin{equation}
    \mathrm{Seasonality}(x)
    =
    \max\left(0, 1-\frac{\mathrm{Var}(R)}{\mathrm{Var}(x-G)}\right).
    \label{eq:no-cpcd-seasonality}
\end{equation}
Dataset-level seasonality is obtained by averaging valid turbine-level values.

\noindent\textbf{Shifting.}
Shifting describes how strongly the distribution of a target power series drifts over time. In wind power data, such drift may be caused by changing weather regimes, turbine operating states, curtailment, or abnormal events. Larger values indicate stronger deviation from a stable temporal distribution. Algorithm~\ref{alg:no-cpcd-shifting} summarizes the threshold-exceedance procedure used to compute this descriptor.

\begin{algorithm}[t]
\caption{Calculating shifting for a target power series.}
\label{alg:no-cpcd-shifting}
\footnotesize
\begin{algorithmic}[1]
\REQUIRE Target series $x\in\mathbb{R}^{T}$; number of thresholds $m$
\ENSURE $\mathrm{Shifting}(x)$
\STATE Normalize $x$ by z-score transformation to obtain $z$
\STATE $z_{\min}\leftarrow\min(z)$, $z_{\max}\leftarrow\max(z)$
\STATE $\mathcal{S}\leftarrow\{z_{\min}+(i-1)(z_{\max}-z_{\min})/m\mid 1\leq i\leq m\}$
\FOR{each threshold $s_i\in\mathcal{S}$}
    \STATE $K_i\leftarrow\{j\mid z_j>s_i,\;1\leq j\leq T\}$
    \STATE $M_i\leftarrow\mathrm{median}(K_i)$
\ENDFOR
\STATE $M'\leftarrow\mathrm{MinMaxNormalize}(M_1,\ldots,M_m)$
\RETURN $\mathrm{Shifting}(x)\leftarrow |\mathrm{median}(M')|$
\end{algorithmic}
\end{algorithm}

\noindent\textbf{Short-term and long-term JSD.}
WPBench further measures distributional irregularity with Jensen--Shannon divergence (JSD). Short-window JSD captures local non-Gaussian fluctuations and abrupt irregularity, while long-window JSD captures broader regime-level distributional deviation. For a window size $s$, let $\mathcal{W}_{s}(x)$ be the set of non-overlapping complete windows from $x$. For each window $w\in\mathcal{W}_{s}(x)$, an empirical density $P_w$ is estimated with a Stone-rule histogram, and a Gaussian reference $Q_w$ is fitted using the same window's mean and standard deviation. The window-level divergence is
\begin{equation}
    \begin{aligned}
    \mathrm{JSD}(P_w,Q_w)
    &=
    \frac{1}{2}\mathrm{KL}(P_w\Vert M_w)
    +
    \frac{1}{2}\mathrm{KL}(Q_w\Vert M_w), \\
    M_w
    &=
    \frac{1}{2}(P_w+Q_w).
    \end{aligned}
    \label{eq:no-cpcd-jsd}
\end{equation}
The score for window size $s$ is
\begin{equation}
    \mathrm{JSD}_{s}(x)
    =
    \frac{1}{|\mathcal{W}_{s}(x)|}
    \sum_{w\in\mathcal{W}_{s}(x)}
    \mathrm{JSD}(P_w,Q_w).
    \label{eq:no-cpcd-jsd-avg}
\end{equation}
WPBench reports short-term and long-term JSD using $s=30$ and $s=336$, respectively, and averages the resulting scores over valid turbine targets.

\subsubsection{Variable Dependency Characteristics}
Variable dependency metrics describe how strongly the target power series is associated with auxiliary variables.

\noindent\textbf{VarCorr.}
For a turbine with target $y$ and covariates $Z=[z_1,\ldots,z_{V-1}]$, VarCorr measures the average absolute target-covariate correlation over the full aligned series:
\begin{equation}
    \mathrm{VarCorr}(y,Z)
    =
    \frac{1}{V-1}
    \sum_{v=1}^{V-1}
    |\rho(y,z_v)|.
    \label{eq:no-cpcd-var-corr}
\end{equation}

\noindent\textbf{Variable TGV and GSD.}
Static correlation alone cannot show whether the role of covariates changes over time. WPBench therefore builds a dynamic target-covariate graph over non-overlapping windows of length $s=576$. For each window $w$, let $y_w$ and $z_{v,w}$ denote the target and the $v$-th covariate restricted to that window. The target-covariate edge weights are defined by
\begin{equation}
    a_{w,v}
    =
    \operatorname{cos}(y_w,z_{v,w}),
    \qquad
    v=1,\ldots,V-1,
    \label{eq:no-cpcd-var-aw}
\end{equation}
where $\operatorname{cos}(\cdot,\cdot)$ denotes cosine similarity, matching the implementation of the dynamic descriptors. Let $\mathbf{a}_w=[a_{w,1},\ldots,a_{w,V-1}]^\top$. These weights define a symmetric star-shaped adjacency matrix $A^{\mathrm{var}}_w\in\mathbb{R}^{V\times V}$:
\begin{equation}
    (A^{\mathrm{var}}_w)_{1,v+1}
    =
    (A^{\mathrm{var}}_w)_{v+1,1}
    =
    a_{w,v},
    \quad
    v=1,\ldots,V-1,
    \label{eq:no-cpcd-var-adj}
\end{equation}
with zero diagonal entries and zero covariate-covariate entries. The corresponding unnormalized graph Laplacian is
\begin{equation}
    L^{\mathrm{var}}_w
    =
    D^{\mathrm{var}}_w - A^{\mathrm{var}}_w,
    \qquad
    D^{\mathrm{var}}_w
    =
    \mathrm{diag}(A^{\mathrm{var}}_w\mathbf{1}).
    \label{eq:no-cpcd-var-lap}
\end{equation}
Let $\lambda(L^{\mathrm{var}}_w)$ be the sorted real eigenvalue vector. With $K$ complete windows, the dynamic variable-dependency descriptors are
\begin{equation}
    \mathrm{Var\mbox{-}TGV}
    =
    \frac{1}{K-1}
    \sum_{w=1}^{K-1}
    \frac{\|\mathbf{a}_{w+1}-\mathbf{a}_w\|_2}{\sqrt{V-1}},
    \label{eq:no-cpcd-var-tgv}
\end{equation}
\begin{equation}
    \mathrm{Var\mbox{-}GSD}
    =
    \frac{1}{K-1}
    \sum_{w=1}^{K-1}
    \frac{\|\lambda(L^{\mathrm{var}}_{w+1})-\lambda(L^{\mathrm{var}}_w)\|_2}{V}.
    \label{eq:no-cpcd-var-gsd}
\end{equation}
Var-TGV measures how rapidly target-covariate edge weights change, while Var-GSD measures structural change through the Laplacian spectrum. Dataset-level values are averaged over turbines with valid covariates and at least two complete windows.

\subsubsection{Spatial Dependency Characteristics}
Spatial dependency metrics describe relationships among turbine target power series.

\noindent\textbf{SpatialCorr.}
For a multi-turbine dataset, let $\mathbf{Y}=[y_1,\ldots,y_N]$ collect the $N$ turbine targets. SpatialCorr measures the average absolute Pearson correlation over all turbine pairs:
\begin{equation}
    \mathrm{SpatialCorr}(\mathbf{Y})
    =
    \frac{2}{N(N-1)}
    \sum_{1\leq i<j\leq N}
    |\rho(y_i,y_j)|.
    \label{eq:no-cpcd-spatial-corr}
\end{equation}

\noindent\textbf{Spatial TGV and GSD.}
To characterize time-varying spatial interactions, WPBench constructs a turbine graph in each non-overlapping window of length $s=576$. Let $y_{i,w}$ denote the target series of turbine $i$ restricted to window $w$. The spatial adjacency matrix $A^{\mathrm{spa}}_w\in\mathbb{R}^{N\times N}$ is defined elementwise as
\begin{equation}
    (A^{\mathrm{spa}}_w)_{ij}
    =
    \operatorname{cos}(y_{i,w},y_{j,w}),
    \quad
    i\neq j,
    \qquad
    (A^{\mathrm{spa}}_w)_{ii}=1,
    \label{eq:no-cpcd-spatial-aw}
\end{equation}
which follows the cosine-similarity matrix used in the implementation. With
\begin{equation}
    L^{\mathrm{spa}}_w
    =
    D^{\mathrm{spa}}_w - A^{\mathrm{spa}}_w,
    \qquad
    D^{\mathrm{spa}}_w
    =
    \mathrm{diag}(A^{\mathrm{spa}}_w\mathbf{1}),
    \label{eq:no-cpcd-spatial-lap}
\end{equation}
the spatial dynamic descriptors are
\begin{equation}
    \mathrm{Spatial\mbox{-}TGV}
    =
    \frac{1}{K-1}
    \sum_{w=1}^{K-1}
    \frac{\|A^{\mathrm{spa}}_{w+1}-A^{\mathrm{spa}}_w\|_{F}}{N},
    \label{eq:no-cpcd-spatial-tgv}
\end{equation}
\begin{equation}
    \mathrm{Spatial\mbox{-}GSD}
    =
    \frac{1}{K-1}
    \sum_{w=1}^{K-1}
    \frac{\|\lambda(L^{\mathrm{spa}}_{w+1})-\lambda(L^{\mathrm{spa}}_w)\|_2}{N\sqrt{N}}.
    \label{eq:no-cpcd-spatial-gsd}
\end{equation}
Spatial-TGV measures volatility in turbine-to-turbine edge weights, and Spatial-GSD measures spectral changes in the turbine interaction graph. The general TGV and GSD computations are summarized in Algorithms~\ref{alg:no-cpcd-tgv} and~\ref{alg:no-cpcd-gsd}.

\begin{algorithm}[t]
\caption{Calculating temporal graph volatility (TGV).}
\label{alg:no-cpcd-tgv}
\footnotesize
\begin{algorithmic}[1]
\REQUIRE Window-level dependency representations $\{G_w\}_{w=1}^{K}$; normalization constant $d_{\mathrm{TGV}}$
\ENSURE $\mathrm{TGV}$
\STATE $\mathcal{V}\leftarrow\emptyset$
\FOR{$w=1$ to $K-1$}
    \STATE $v_w\leftarrow\|G_{w+1}-G_w\|/d_{\mathrm{TGV}}$
    \STATE $\mathcal{V}\leftarrow\mathcal{V}\cup\{v_w\}$
\ENDFOR
\RETURN $\mathrm{TGV}\leftarrow \mathrm{mean}(\mathcal{V})$
\end{algorithmic}
\end{algorithm}

\begin{algorithm}[t]
\caption{Calculating graph spectral divergence (GSD).}
\label{alg:no-cpcd-gsd}
\footnotesize
\begin{algorithmic}[1]
\REQUIRE Window-level adjacency matrices $\{A_w\}_{w=1}^{K}$; normalization constant $d_{\mathrm{GSD}}$
\ENSURE $\mathrm{GSD}$
\FOR{$w=1$ to $K$}
    \STATE $L_w\leftarrow\mathrm{diag}(A_w\mathbf{1})-A_w$
    \STATE $\ell_w\leftarrow$ sorted eigenvalues of $L_w$
\ENDFOR
\STATE $\mathcal{D}\leftarrow\emptyset$
\FOR{$w=1$ to $K-1$}
    \STATE $d_w\leftarrow\|\ell_{w+1}-\ell_w\|_2/d_{\mathrm{GSD}}$
    \STATE $\mathcal{D}\leftarrow\mathcal{D}\cup\{d_w\}$
\ENDFOR
\RETURN $\mathrm{GSD}\leftarrow \mathrm{mean}(\mathcal{D})$
\end{algorithmic}
\end{algorithm}

\section{WPBench}
\begin{figure*}[t]
\centering
\includegraphics[width=\textwidth]{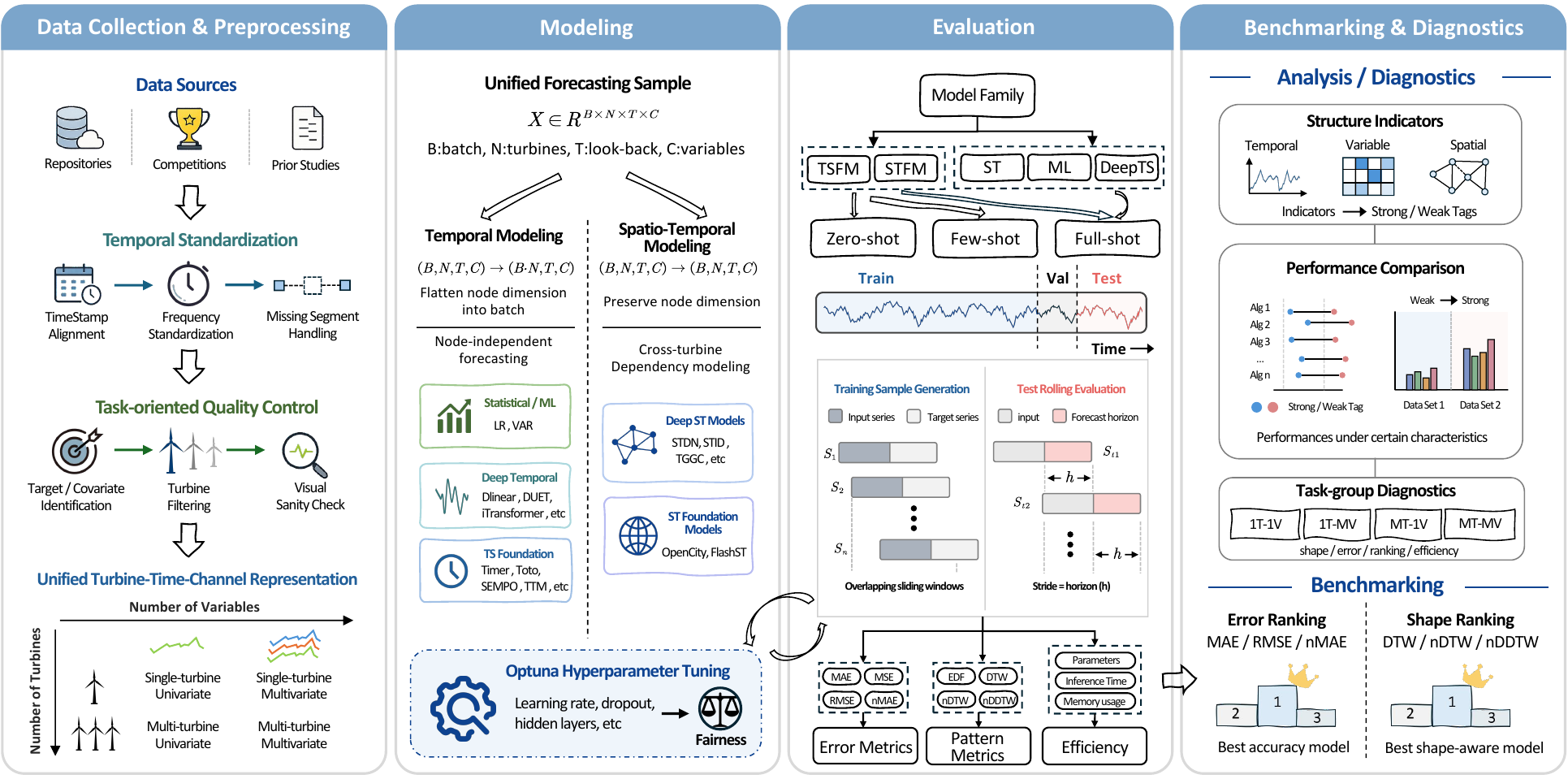}
\caption{Overview of the WPBench benchmark workflow. WPBench transforms heterogeneous public wind power datasets into structure-aware forecasting tasks, evaluates representative model families, and supports multi-dimensional performance and diagnostic analysis.}
\label{fig:wpbench_workflow}
\end{figure*}

Figure~\ref{fig:wpbench_workflow} presents the overall workflow of WPBench, covering data collection and preprocessing, model organization, unified evaluation, and benchmarking with structure-aware diagnostics. The following subsections describe these components in detail.

\subsection{Dataset Collection and Preprocessing}

WPBench contains 26 publicly accessible wind power datasets collected from public repositories, forecasting competitions, and datasets released with prior studies. These datasets differ in sampling frequency, variable composition, turbine scale, and data quality. To support unified modeling and evaluation, we standardize them into a turbine--time--channel representation, where each dataset contains one or more target power series at the turbine or site level and, when available, auxiliary covariates such as SCADA variables or meteorological measurements.

The preprocessing pipeline follows a consistent but minimally intrusive strategy. We align timestamps and sampling frequencies, check temporal continuity, handle missing or unreliable segments, identify the target wind power channel, and retain semantically meaningful covariates. For multi-turbine datasets, we further verify turbine alignment and remove turbine records with severe missingness, long constant periods, abnormal jumps, or clear temporal inconsistency. Finally, visual sanity checks are conducted to detect artifacts that are difficult to capture by rules alone. This process removes obvious data artifacts while preserving valid temporal variation, covariate information, and turbine-level structure.
\subsection{Wind-Specific Dataset Organization}
\label{sec:dataset_taxonomy}

Wind power datasets differ not only in scale but also in the structure of the forecasting problem. WPBench organizes datasets along two wind-specific axes: the number of turbines and the number of observed variables. The turbine axis determines whether the task contains spatial dependencies, while the variable axis determines whether the model can exploit target-covariate relationships. This yields four task types: single-turbine univariate (1T1V), single-turbine multivariate (1TMV), multi-turbine univariate (MT1V), and multi-turbine multivariate (MTMV). Table~\ref{tab:wpbench_dataset_statistics} summarizes the 26 datasets under this organization. All datasets are chronologically split into training, validation, and testing sets according to the ratios indicated in the Split column (e.g., 7:1:2).

\begin{table}[t]
\centering
\caption{Statistics of WPBench datasets.}
\label{tab:wpbench_dataset_statistics}
\scriptsize
\setlength{\tabcolsep}{6.0pt}
\renewcommand{\arraystretch}{0.92}
\resizebox{\columnwidth}{!}{
\begin{tabular}{@{}lccccc@{}}
\toprule
Dataset & Frequency & Length & $N$ & $C$ & Split \\
\midrule
\multicolumn{6}{@{}l}{\emph{Single-turbine univariate datasets}} \\
Wind Farm A~\cite{guuck2024care} & 10mins & 72,158 & 1 & 1 & 7:1:2 \\
Wind Farm B~\cite{guuck2024care} & 10mins & 22,615 & 1 & 1 & 7:1:2 \\
Chalmers~\cite{fogelstrom2023bjorko} & 1min & 142,761 & 1 & 1 & 7:1:2 \\
Maelstrom 1~\cite{climetlab_plugin_a6} & 10mins & 239,028 & 1 & 1 & 7:1:2 \\
Maelstrom 2~\cite{climetlab_plugin_a6} & 10mins & 239,019 & 1 & 1 & 7:1:2 \\
Maelstrom 3~\cite{climetlab_plugin_a6} & 10mins & 239,028 & 1 & 1 & 7:1:2 \\
Maelstrom 4~\cite{climetlab_plugin_a6} & 10mins & 239,028 & 1 & 1 & 7:1:2 \\
\midrule
\multicolumn{6}{@{}l}{\emph{Single-turbine multivariate datasets}} \\
Yalova~\cite{effenberger2022yalova} & 10mins & 52,560 & 1 & 4 & 7:1:2 \\
Wind Farm Site 1~\cite{chen2022solar} & 15mins & 70,176 & 1 & 12 & 7:1:2 \\
Wind Farm Site 2~\cite{chen2022solar} & 15mins & 70,176 & 1 & 11 & 7:1:2 \\
Wind Farm Site 3~\cite{chen2022solar} & 15mins & 70,176 & 1 & 12 & 7:1:2 \\
Wind Farm Site 4~\cite{chen2022solar} & 15mins & 70,176 & 1 & 12 & 7:1:2 \\
Wind Farm Site 6~\cite{chen2022solar} & 15mins & 70,176 & 1 & 12 & 7:1:2 \\
Kaggle1~\cite{kaggle_wind_power_forecasting} & 10mins & 15,984 & 1 & 19 & 7:1:2 \\
SCADA\_Fault~\cite{kaggle_iiot_wind_turbine} & 10mins & 47,427 & 1 & 64 & 7:1:2 \\
\midrule
\multicolumn{6}{@{}l}{\emph{Multi-turbine univariate datasets}} \\
Wind Farm C~\cite{guuck2024care} & 10mins & 10,007 & 21 & 1 & 7:1:2 \\
AEMO~\cite{dowell2015very} & 5mins & 210,479 & 22 & 1 & 7:1:2 \\
COSMO~\cite{jensen2017reeurope} & 1hour & 26,304 & 38 & 1 & 7:1:2 \\
ECMWF~\cite{jensen2017reeurope} & 1hour & 26,304 & 38 & 1 & 7:1:2 \\
\midrule
\multicolumn{6}{@{}l}{\emph{Multi-turbine multivariate datasets}} \\
GEFCom2012~\cite{hong2014gefcom2012} & 1hour & 26,244 & 7 & 5 & 7:1:2 \\
GEFCom2014~\cite{hong2016gefcom2014} & 1hour & 17,538 & 10 & 5 & 7:1:2 \\
Kelmarsh~\cite{plumley2022kelmarsh} & 10mins & 197,977 & 6 & 3 & 7:1:2 \\
Penmanshiel~\cite{plumley2022penmanshiel} & 10mins & 245,232 & 14 & 3 & 7:1:2 \\
SDWPF~\cite{zhou2024sdwpf} & 10mins & 35,280 & 134 & 10 & 7:1:2 \\
UEPS~\cite{passos2019coastal} & 10mins & 52,560 & 20 & 3 & 7:1:2 \\
UEBB~\cite{passos2019coastal} & 10mins & 52,560 & 32 & 3 & 7:1:2 \\
\bottomrule
\end{tabular}}
\end{table}

Compared with existing wind forecasting resources, WPBench emphasizes systematic coverage of wind-data organizations. Existing public resources are usually concentrated on specific representative scenarios: GEFCom focuses mainly on competition-style wind power forecasting~\cite{hong2014gefcom2012,hong2016gefcom2014}, SDWPF emphasizes multi-turbine wind farm forecasting~\cite{zhou2024sdwpf}, and OpenWPF~\cite{xu2025cross} mainly covers multivariate and multi-turbine forecasting datasets. In contrast, WPBench covers single-turbine univariate, single-turbine multivariate, multi-turbine univariate, and multi-turbine multivariate tasks, enabling unified evaluation under target-series modeling, covariate-aware modeling, spatial dependency modeling, and their combined setting.

\begin{figure}[h]
\centering
\includegraphics[width=0.4\textwidth]{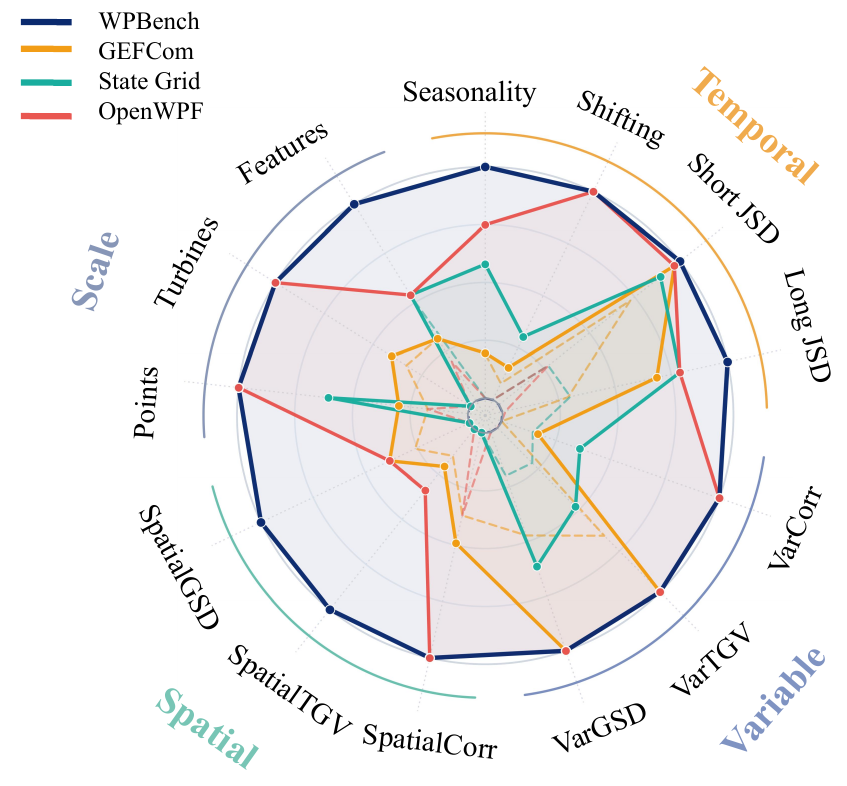}
\caption{Dataset-characteristic coverage of WPBench and representative wind forecasting resources. The radar chart compares normalized coverage profiles across temporal patterns, variable dependencies, spatial dependencies, and data scale. Solid contours denote group-wise maximum values, while dashed contours denote group-wise minimum values.}
\label{fig:dataset_coverage_distribution}
\end{figure}

Figure~\ref{fig:dataset_coverage_distribution} further compares WPBench with representative wind forecasting resources in terms of temporal patterns, variable relationships, spatial dependencies, and data scale. Compared with existing resources, WPBench provides a broader and more balanced coverage envelope, with stronger coverage of dataset characteristics involving complex variable relationships, spatial dependencies, and scale variations.

\subsection{Forecasting Models}
To compare different forecasting paradigms in WPF, WPBench includes 19 representative models from five categories: 1) statistical and machine learning models: LR\cite{herzen2022darts,kedem2005regression} and VAR\cite{toda1994var}; 2) deep temporal models: DLinear\cite{zeng2023dlinear}, XLinear\cite{chen2026xlinear}, DUET\cite{qiu2025duet}, PatchTST\cite{nie2022patchtst}, iTransformer\cite{liu2024itransformer}, and xPatch\cite{stitsyuk2025xpatch}; 3) time-series foundation models: Timer\cite{liu2024timer}, TTM\cite{ekambaram2024ttm}, TOTO\cite{cohen2026TOTO}, and SEMPO\cite{he2026sempo}; 4) deep spatio-temporal models: STID\cite{shao2022stid}, STDN\cite{cao2025stdn}, TGGC\cite{jin2025TGGC}, and BigST\cite{Han2024BigST}; and 5) spatio-temporal foundation models: OpenCity\cite{Li2024Opencity}, FactoST-UTP\cite{Zhong2025Factost}, and FactoST-STA\cite{Zhong2025Factost}. For foundation models, WPBench further provides three adaptation settings: zero-shot, few-shot, and full-shot.

\subsection{Evaluator}

\noindent\textbf{Forecasting Protocol.}
WPBench evaluates models under a fixed-window multi-step forecasting protocol. Given a look-back window of $L=576$, each model predicts $H\in\{12,24,72,144\}$ future steps, covering short-term and longer-horizon settings. Sliding windows are constructed from the training split, while testing follows a non-overlapping rolling protocol where each test window advances by $H$ steps. All models are compared on the same splits, dataset-horizon units, and rolling windows. For fair comparison, key hyperparameters of trainable models are tuned on the validation split using Optuna\cite{akiba2019optuna}, and the best validation configuration is fixed for test evaluation. The experiments are implemented in PyTorch 2.6.0+cu124\cite{paszke2019pytorch} with Python 3.11 and run on NVIDIA A800 80GB GPUs.

\noindent\textbf{Evaluation Metrics.}
WPBench evaluates numerical accuracy and forecast-curve fidelity. We use six error metrics, including MAE, MSE, nMAE, nMSE, nRMSE, and nMBE, together with four shape-aware metrics, including DTW, nDTW, nDDTW, and EDF. For one testing window, let $e_{i,k}=F_{i,k}-Y_{i,k}$ and $\tilde{e}_{i,k}=e_{i,k}/C_i$, where $Y_{i,k}$ and $F_{i,k}$ are the ground-truth and forecast target values of turbine or site $i$ at horizon step $k$, and $C_i$ is the rated capacity or a capacity-like normalizer of the corresponding target when available; otherwise, it is approximated by the maximum target value in the training split. The error metrics are defined as follows:
\[
\begin{aligned}
\mathit{MAE} &= \frac{1}{NH}\sum_{i=1}^{N}\sum_{k=1}^{H}|e_{i,k}|, &
\mathit{MSE} &= \frac{1}{NH}\sum_{i=1}^{N}\sum_{k=1}^{H}e_{i,k}^{2}, \\
\mathit{nMAE} &= \frac{1}{NH}\sum_{i=1}^{N}\sum_{k=1}^{H}|\tilde{e}_{i,k}|, &
\mathit{nMSE} &= \frac{1}{NH}\sum_{i=1}^{N}\sum_{k=1}^{H}\tilde{e}_{i,k}^{2}, \\
\mathit{nRMSE} &= \sqrt{\mathit{nMSE}}, &
\mathit{nMBE} &= \frac{1}{NH}\sum_{i=1}^{N}\sum_{k=1}^{H}\tilde{e}_{i,k}.
\end{aligned}
\]

\noindent
For forecast-curve fidelity, we show the core DTW alignment cost and the EDF definition:
\[
\begin{aligned}
\mathrm{DTW}(Y_i,F_i)
&= \min_{\pi_i}\sum_{(p,q)\in\pi_i} d(Y_{i,p},F_{i,q}), \\
\mathit{EDF}
&= \frac{\mathit{nRMSE}}{\mathit{nMAE}},
\end{aligned}
\]

\noindent
where $N$ is the number of turbines or site-level target series, $H$ is the forecasting horizon, $Y_i$ and $F_i$ denote the corresponding ground-truth and forecast target curves of turbine or site $i$, $\pi_i$ is a warping path, $(p,q)$ denotes an aligned pair of horizon indices between $Y_i$ and $F_i$ along $\pi_i$, and $d(\cdot,\cdot)$ is the point-wise distance. Dataset-level DTW is computed by averaging $\mathrm{DTW}(Y_i,F_i)$ over $i$. The normalized DTW, nDTW, is computed by replacing $Y_i$ and $F_i$ with capacity-normalized curves $\bar{Y}_i=Y_i/C_i$ and $\bar{F}_i=F_i/C_i$, while nDDTW applies DTW to the first-order change sequences $\Delta\bar{Y}_i$ and $\Delta\bar{F}_i$. EDF characterizes the dispersion pattern of forecasting errors: a larger EDF indicates that large deviations contribute more strongly to the overall error, while an EDF closer to 1 indicates a more uniform error distribution. For single-turbine datasets, $N=1$. In the main experiments, nMAE/nRMSE and nDTW/nDDTW are used as the primary accuracy and shape-aware metrics, respectively.    

\section{Experiments}

Rather than treating WPBench as a static leaderboard, our experiments use it as a diagnostic instrument to study how model families respond to wind-specific temporal patterns, target-covariate dependencies, and turbine-level spatial structures. We first report overall performance across the four wind data structures, and then use the characterization metrics defined in the preliminaries to analyze when different modeling assumptions hold or fail.

\subsection{Overall Results under Scope-aware Ranking}

WPBench covers four forecasting settings: single-turbine
univariate (1T1V), single-turbine multivariate (1TMV), multi-
turbine univariate (MT1V), and multi-turbine multivariate
(MTMV). Because model families differ in structural applicability,
we report detailed nMAE results separately for single-turbine and
multi-turbine datasets. Tables~\ref{tab:ch5_nmae_detailed_single}
and~\ref{tab:ch5_nmae_detailed_mt} show the nMAE results for
these two groups; models such as VAR have missing entries where
the corresponding data structure is not applicable. Each dataset is
reported with two horizon groups: Short averages horizons 12 and
24, while Long averages horizons 72 and 144.

\begin{table*}[t]
\centering
\caption{nMAE detailed results on single-turbine datasets.}
\label{tab:ch5_nmae_detailed_single}
\scriptsize
\setlength{\tabcolsep}{5pt}
\renewcommand{\arraystretch}{0.88}
\resizebox{1\textwidth}{!}{%
\begin{tabular}{@{}c|l|c|cccccc|cc|cccc|ccc|c@{}}
\toprule
\multicolumn{1}{c|}{\multirow{2}{*}{Type}}
& \multicolumn{1}{c|}{\multirow{2}{*}{Dataset}}
& Horizon
& \multicolumn{1}{|c}{DLinear}
& \multicolumn{1}{c}{XLinear}
& \multicolumn{1}{c}{DUET}
& \multicolumn{1}{c}{Patch}
& \multicolumn{1}{c}{iTrans}
& \multicolumn{1}{c|}{xPatch}
& \multicolumn{1}{c}{LR}
& \multicolumn{1}{c|}{VAR}
& \multicolumn{1}{c}{Timer}
& \multicolumn{1}{c}{TTM}
& \multicolumn{1}{c}{TOTO}
& \multicolumn{1}{c|}{SEMPO}
& \multicolumn{1}{c}{STID}
& \multicolumn{1}{c}{TGGC}
& \multicolumn{1}{c|}{BigST}
& \multicolumn{1}{c}{FUTP} \\
\cmidrule(lr){4-19}
& & & nMAE & nMAE & nMAE & nMAE & nMAE & nMAE & nMAE & nMAE & nMAE & nMAE & nMAE & nMAE & nMAE & nMAE & nMAE & nMAE \\
\midrule
\multirow{14}{*}{\rotatebox[origin=c]{90}{\emph{1T1V}}}
& \multicolumn{1}{c|}{\multirow{2}{*}{Wind Farm A}} & Short & 0.096 & 0.095 & 0.100 & 0.107 & 0.104 & 0.089 & 0.096 & -- & 0.101 & 0.091 & \best{0.081} & 0.100 & 0.102 & 0.110 & 0.107 & \secondbest{0.082} \\
& & Long & 0.179 & 0.168 & \secondbest{0.167} & 0.183 & 0.178 & 0.171 & 0.179 & -- & 0.212 & 0.173 & \best{0.167} & 0.188 & 0.185 & 0.195 & 0.214 & 0.170 \\
\addlinespace[1pt]\cline{2-19}\addlinespace[1pt]
& \multicolumn{1}{c|}{\multirow{2}{*}{Wind Farm B}} & Short & 0.098 & 0.106 & 0.105 & 0.130 & 0.124 & 0.110 & 0.095 & -- & 0.108 & 0.100 & \secondbest{0.093} & 0.112 & 0.116 & 0.114 & 0.137 & \best{0.093} \\
& & Long & 0.170 & 0.173 & 0.179 & 0.186 & 0.190 & 0.206 & \best{0.165} & -- & 0.224 & 0.184 & 0.178 & 0.186 & 0.173 & 0.182 & 0.226 & \secondbest{0.170} \\
\addlinespace[1pt]\cline{2-19}\addlinespace[1pt]
& \multicolumn{1}{c|}{\multirow{2}{*}{Chalmers}} & Short & 0.003 & 0.002 & 0.001 & 0.003 & 0.004 & 0.001 & 0.004 & -- & 0.002 & 0.001 & \secondbest{0.001} & 0.002 & 0.005 & 0.006 & 0.722 & \best{0.001} \\
& & Long & 0.011 & 0.004 & 0.004 & \best{0.004} & 0.004 & \secondbest{0.004} & 0.014 & -- & 0.005 & 0.008 & 0.005 & 0.004 & 0.013 & 0.008 & 0.568 & 0.005 \\
\addlinespace[1pt]\cline{2-19}\addlinespace[1pt]
& \multicolumn{1}{c|}{\multirow{2}{*}{Maelstrom 1}} & Short & 0.069 & 0.068 & 0.072 & 0.086 & 0.076 & 0.069 & 0.069 & -- & 0.071 & 0.075 & \secondbest{0.067} & 0.072 & 0.072 & 0.070 & 0.072 & \best{0.065} \\
& & Long & 0.117 & 0.114 & \secondbest{0.111} & 0.120 & 0.117 & 0.120 & 0.118 & -- & 0.133 & 0.115 & \best{0.110} & 0.125 & 0.118 & 0.124 & 0.121 & 0.117 \\
\addlinespace[1pt]\cline{2-19}\addlinespace[1pt]
& \multicolumn{1}{c|}{\multirow{2}{*}{Maelstrom 2}} & Short & 0.060 & 0.059 & 0.060 & 0.071 & 0.065 & 0.058 & 0.060 & -- & 0.059 & 0.061 & \secondbest{0.057} & 0.060 & 0.059 & 0.060 & 0.061 & \best{0.056} \\
& & Long & 0.093 & 0.088 & \best{0.086} & 0.095 & 0.090 & 0.092 & 0.094 & -- & 0.099 & 0.089 & \secondbest{0.087} & 0.095 & 0.094 & 0.094 & 0.098 & 0.093 \\
\addlinespace[1pt]\cline{2-19}\addlinespace[1pt]
& \multicolumn{1}{c|}{\multirow{2}{*}{Maelstrom 3}} & Short & 0.078 & 0.076 & 0.080 & 0.085 & 0.084 & 0.076 & 0.077 & -- & 0.076 & 0.078 & \secondbest{0.075} & 0.080 & 0.077 & 0.079 & 0.079 & \best{0.074} \\
& & Long & 0.123 & 0.121 & \secondbest{0.119} & 0.130 & 0.130 & 0.122 & 0.123 & -- & 0.138 & 0.123 & \best{0.118} & 0.131 & 0.125 & 0.127 & 0.127 & 0.127 \\
\addlinespace[1pt]\cline{2-19}\addlinespace[1pt]
& \multicolumn{1}{c|}{\multirow{2}{*}{Maelstrom 4}} & Short & 0.076 & \secondbest{0.074} & 0.078 & 0.081 & 0.081 & 0.075 & 0.076 & -- & 0.076 & 0.078 & 0.074 & 0.078 & 0.079 & 0.077 & 0.078 & \best{0.072} \\
& & Long & 0.126 & 0.123 & \secondbest{0.122} & 0.134 & 0.129 & 0.126 & 0.128 & -- & 0.144 & 0.123 & \best{0.121} & 0.136 & 0.127 & 0.133 & 0.135 & 0.132 \\
\midrule
\multirow{16}{*}{\rotatebox[origin=c]{90}{\emph{1TMV}}}
& \multicolumn{1}{c|}{\multirow{2}{*}{Wind Farm Site 1}} & Short & 0.096 & 0.096 & 0.129 & 0.109 & 0.117 & 0.098 & 0.096 & 0.095 & 0.103 & 0.099 & \secondbest{0.092} & 0.106 & 0.101 & 0.108 & 0.118 & \best{0.090} \\
& & Long & 0.172 & 0.170 & 0.182 & 0.186 & 0.191 & 0.187 & 0.167 & \best{0.167} & 0.201 & 0.173 & \secondbest{0.167} & 0.185 & 0.174 & 0.175 & 0.188 & 0.191 \\
\addlinespace[1pt]\cline{2-19}\addlinespace[1pt]
& \multicolumn{1}{c|}{\multirow{2}{*}{Wind Farm Site 2}} & Short & 0.108 & 0.107 & 0.160 & 0.128 & 0.126 & 0.109 & 0.105 & 0.104 & 0.112 & 0.107 & \secondbest{0.095} & 0.116 & 0.120 & 0.115 & 0.136 & \best{0.094} \\
& & Long & 0.209 & 0.212 & 0.233 & 0.220 & 0.222 & 0.218 & \best{0.198} & 0.209 & 0.245 & 0.213 & \secondbest{0.203} & 0.220 & 0.214 & 0.214 & 0.235 & 0.237 \\
\addlinespace[1pt]\cline{2-19}\addlinespace[1pt]
& \multicolumn{1}{c|}{\multirow{2}{*}{Wind Farm Site 3}} & Short & 0.067 & 0.066 & 0.098 & 0.083 & 0.078 & 0.067 & 0.069 & 0.068 & 0.064 & 0.064 & \secondbest{0.061} & 0.070 & 0.075 & 0.073 & 0.104 & \best{0.059} \\
& & Long & 0.133 & 0.125 & 0.125 & 0.131 & 0.129 & 0.126 & 0.143 & 0.133 & 0.137 & 0.128 & \best{0.118} & 0.136 & 0.138 & 0.139 & 0.173 & \secondbest{0.118} \\
\addlinespace[1pt]\cline{2-19}\addlinespace[1pt]
& \multicolumn{1}{c|}{\multirow{2}{*}{Wind Farm Site 4}} & Short & 0.073 & 0.076 & 0.124 & 0.102 & 0.102 & 0.074 & 0.079 & 0.074 & 0.079 & 0.073 & \best{0.068} & 0.089 & 0.083 & 0.081 & 0.102 & \secondbest{0.068} \\
& & Long & 0.185 & 0.185 & 0.211 & 0.216 & 0.202 & 0.198 & 0.189 & \secondbest{0.181} & 0.247 & 0.182 & \best{0.172} & 0.207 & 0.195 & 0.191 & 0.240 & 0.190 \\
\addlinespace[1pt]\cline{2-19}\addlinespace[1pt]
& \multicolumn{1}{c|}{\multirow{2}{*}{Wind Farm Site 6}} & Short & 0.062 & 0.059 & 0.095 & 0.070 & 0.071 & 0.059 & 0.096 & 0.057 & 0.059 & 0.057 & \best{0.053} & 0.061 & 0.064 & 0.072 & 0.176 & \secondbest{0.056} \\
& & Long & 0.145 & \best{0.107} & 0.125 & 0.140 & 0.139 & 0.136 & 0.201 & 0.127 & 0.154 & 0.135 & \secondbest{0.110} & 0.135 & 0.169 & 0.153 & 0.278 & 0.160 \\
\addlinespace[1pt]\cline{2-19}\addlinespace[1pt]
& \multicolumn{1}{c|}{\multirow{2}{*}{Kaggle1}} & Short & \secondbest{0.109} & 0.116 & 0.136 & 0.133 & 0.137 & 0.118 & 0.123 & 0.119 & 0.120 & 0.109 & \best{0.108} & 0.122 & 0.118 & 0.126 & 0.135 & 0.113 \\
& & Long & \best{0.152} & 0.160 & 0.178 & 0.174 & 0.183 & 0.177 & 0.167 & 0.157 & 0.212 & \secondbest{0.152} & 0.155 & 0.171 & 0.161 & 0.166 & 0.165 & 0.168 \\
\addlinespace[1pt]\cline{2-19}\addlinespace[1pt]
& \multicolumn{1}{c|}{\multirow{2}{*}{SCADA\_Fault}} & Short & 0.121 & 0.124 & 0.148 & 0.139 & 0.150 & 0.125 & 0.141 & 0.125 & 0.128 & 0.120 & \secondbest{0.114} & 0.129 & 0.129 & 0.130 & 0.191 & \best{0.113} \\
& & Long & \secondbest{0.211} & 0.218 & 0.251 & 0.231 & 0.249 & 0.230 & 0.230 & 0.219 & 0.250 & 0.214 & \best{0.197} & 0.234 & 0.215 & 0.226 & 0.503 & 0.241 \\
\addlinespace[1pt]\cline{2-19}\addlinespace[1pt]
& \multicolumn{1}{c|}{\multirow{2}{*}{Yalova}} & Short & 0.107 & 0.111 & 0.140 & 0.133 & 0.143 & 0.111 & 0.105 & 0.104 & 0.116 & 0.105 & \best{0.096} & 0.123 & 0.119 & 0.116 & 0.145 & \secondbest{0.102} \\
& & Long & 0.221 & 0.229 & 0.244 & 0.237 & 0.257 & 0.226 & \secondbest{0.213} & 0.216 & 0.257 & 0.217 & \best{0.207} & 0.251 & 0.226 & 0.226 & 0.262 & 0.222 \\
\bottomrule
\end{tabular}%
}
\vspace{2pt}

\footnotesize
1T1V/1TMV-only nMAE results. Short averages horizons 12 and 24; Long averages horizons 72 and 144. Red bold entries denote the best results, and blue italic underlined entries denote the second-best results in each dataset--horizon-group row; ``--'' indicates unavailable results.
\end{table*}

\begin{table*}[t]
\centering
\caption{nMAE detailed results on multi-turbine datasets.}
\label{tab:ch5_nmae_detailed_mt}
\scriptsize
\setlength{\tabcolsep}{2pt}
\renewcommand{\arraystretch}{0.88}
\resizebox{\textwidth}{!}{%
\begin{tabular}{@{}c|l|c|cccccc|cc|cccc|cccc|c|cc@{}}
\toprule
\multicolumn{1}{c|}{\multirow{2}{*}{Type}}
& \multicolumn{1}{c|}{\multirow{2}{*}{Dataset}}
& Horizon
& \multicolumn{1}{|c}{DLinear}
& \multicolumn{1}{c}{XLinear}
& \multicolumn{1}{c}{DUET}
& \multicolumn{1}{c}{Patch}
& \multicolumn{1}{c}{iTrans}
& \multicolumn{1}{c|}{xPatch}
& \multicolumn{1}{c}{LR}
& \multicolumn{1}{c|}{VAR}
& \multicolumn{1}{c}{Timer}
& \multicolumn{1}{c}{TTM}
& \multicolumn{1}{c}{TOTO}
& \multicolumn{1}{c|}{SEMPO}
& \multicolumn{1}{c}{STID}
& \multicolumn{1}{c}{TGGC}
& \multicolumn{1}{c}{BigST}
& \multicolumn{1}{c|}{STDN}
& \multicolumn{1}{c|}{FUTP}
& \multicolumn{1}{c}{OpenC}
& \multicolumn{1}{c|}{FSTA} \\
\cmidrule(lr){4-22}
& & & nMAE & nMAE & nMAE & nMAE & nMAE & nMAE & nMAE & nMAE & nMAE & nMAE & nMAE & nMAE & nMAE & nMAE & nMAE & nMAE & nMAE & nMAE & nMAE \\
\midrule

\multirow{8}{*}{\rotatebox[origin=c]{90}{\emph{MT1V}}}
& \multicolumn{1}{c|}{\multirow{2}{*}{Wind Farm C}} & Short & 0.105 & 0.110 & 0.112 & 0.140 & 0.126 & 0.108 & 0.105 & -- & 0.114 & 0.101 & \secondbest{0.097} & 0.118 & 0.116 & 0.138 & 0.188 & 0.102 & \best{0.095} & 0.106 & 0.123 \\
& & Long & 0.198 & 0.193 & 0.192 & 0.225 & 0.237 & 0.206 & 0.196 & -- & 0.232 & 0.191 & \secondbest{0.191} & 0.213 & 0.204 & 0.224 & 0.234 & 0.222 & 0.199 & \best{0.187} & 0.262 \\
\addlinespace[1pt]\cline{2-22}\addlinespace[1pt]

& \multicolumn{1}{c|}{\multirow{2}{*}{ECMWF}} & Short & 0.074 & 0.073 & 0.069 & 0.076 & 0.080 & 0.070 & 0.074 & -- & 0.075 & 0.073 & 0.067 & 0.075 & 0.074 & 0.079 & 0.080 & \secondbest{0.065} & \best{0.064} & 0.073 & 0.078 \\
& & Long & 0.104 & 0.099 & 0.093 & 0.103 & 0.107 & 0.098 & 0.106 & -- & 0.107 & 0.100 & \secondbest{0.093} & 0.104 & 0.106 & 0.108 & 0.114 & 0.104 & \best{0.092} & 0.096 & 0.111 \\
\addlinespace[1pt]\cline{2-22}\addlinespace[1pt]

& \multicolumn{1}{c|}{\multirow{2}{*}{COSMO}} & Short & 0.088 & 0.086 & 0.082 & 0.090 & 0.093 & 0.082 & 0.087 & -- & 0.085 & 0.087 & 0.079 & 0.088 & 0.088 & 0.089 & 0.092 & \secondbest{0.077} & \best{0.077} & 0.085 & 0.091 \\
& & Long & 0.119 & 0.114 & 0.105 & 0.118 & 0.120 & 0.109 & 0.120 & -- & 0.125 & 0.116 & \secondbest{0.105} & 0.119 & 0.120 & 0.125 & 0.127 & 0.118 & \best{0.104} & 0.109 & 0.126 \\
\addlinespace[1pt]\cline{2-22}\addlinespace[1pt]

& \multicolumn{1}{c|}{\multirow{2}{*}{AEMO}} & Short & 0.079 & 0.078 & 0.097 & 0.088 & 0.082 & 0.075 & 0.079 & -- & 0.079 & 0.080 & 0.076 & 0.080 & 0.077 & 0.079 & 0.077 & \best{0.072} & \secondbest{0.074} & 0.078 & 0.079 \\
& & Long & 0.175 & 0.175 & 0.180 & 0.183 & 0.177 & 0.171 & 0.176 & -- & 0.196 & 0.172 & 0.174 & 0.204 & 0.171 & 0.167 & 0.169 & \best{0.166} & \secondbest{0.166} & 0.169 & 0.197 \\

\midrule

\multirow{14}{*}{\rotatebox[origin=c]{90}{\emph{MTMV}}}
& \multicolumn{1}{c|}{\multirow{2}{*}{GEFCom2012}} & Short & 0.208 & 0.211 & 0.220 & 0.222 & 0.217 & 0.210 & 0.195 & \best{0.193} & 0.208 & 0.207 & 0.204 & 0.211 & 0.217 & 0.217 & 0.223 & \secondbest{0.195} & 0.199 & 0.209 & 0.218 \\
& & Long & 0.266 & 0.269 & \secondbest{0.263} & 0.268 & 0.275 & 0.274 & 0.264 & \best{0.260} & 0.279 & 0.268 & 0.269 & 0.275 & 0.268 & 0.266 & 0.277 & 0.299 & 0.276 & 0.275 & 0.291 \\
\addlinespace[1pt]\cline{2-22}\addlinespace[1pt]

& \multicolumn{1}{c|}{\multirow{2}{*}{GEFCom2014}} & Short & 0.208 & 0.207 & 0.270 & 0.221 & 0.223 & 0.208 & 0.194 & \best{0.192} & 0.206 & 0.208 & 0.196 & 0.211 & 0.212 & 0.232 & 0.226 & \secondbest{0.193} & 0.196 & 0.211 & 0.210 \\
& & Long & 0.263 & 0.269 & 0.274 & 0.269 & 0.275 & 0.272 & \best{0.254} & \secondbest{0.256} & 0.278 & 0.264 & 0.271 & 0.271 & 0.262 & 0.274 & 0.278 & 0.282 & 0.269 & 0.268 & 0.285 \\
\addlinespace[1pt]\cline{2-22}\addlinespace[1pt]

& \multicolumn{1}{c|}{\multirow{2}{*}{Kelmarsh}} & Short & 0.092 & 0.091 & 0.100 & 0.100 & 0.098 & 0.090 & 0.093 & 0.093 & 0.091 & 0.093 & 0.088 & 0.094 & 0.093 & 0.096 & 0.099 & \secondbest{0.088} & \best{0.086} & 0.092 & 0.093 \\
& & Long & 0.175 & 0.165 & 0.168 & 0.174 & 0.172 & 0.166 & 0.174 & 0.176 & 0.184 & \secondbest{0.163} & \best{0.157} & 0.191 & 0.171 & 0.173 & 0.179 & 0.166 & 0.170 & 0.165 & 0.193 \\
\addlinespace[1pt]\cline{2-22}\addlinespace[1pt]

& \multicolumn{1}{c|}{\multirow{2}{*}{Penmanshiel}} & Short & 0.099 & 0.097 & 0.103 & 0.101 & 0.104 & 0.094 & 0.097 & 0.097 & 0.098 & 0.098 & 0.092 & 0.103 & 0.103 & 0.104 & 0.102 & \best{0.091} & \secondbest{0.091} & 0.096 & 0.103 \\
& & Long & 0.189 & 0.182 & 0.177 & 0.199 & 0.190 & 0.183 & 0.185 & 0.184 & 0.218 & 0.181 & \best{0.170} & 0.202 & 0.188 & 0.189 & 0.192 & \secondbest{0.171} & 0.187 & 0.174 & 0.209 \\
\addlinespace[1pt]\cline{2-22}\addlinespace[1pt]

& \multicolumn{1}{c|}{\multirow{2}{*}{SDWPF}} & Short & 0.095 & 0.092 & 0.098 & 0.106 & 0.118 & 0.088 & 0.145 & 0.091 & 0.093 & 0.085 & \best{0.076} & 0.108 & 0.105 & 0.110 & 0.140 & \secondbest{0.076} & 0.079 & 0.087 & 0.105 \\
& & Long & 0.181 & 0.176 & 0.178 & 0.191 & 0.197 & 0.169 & 0.267 & 0.163 & 0.189 & \secondbest{0.161} & \best{0.149} & 0.180 & 0.197 & 0.197 & 0.197 & 0.189 & 0.176 & 0.172 & 0.208 \\
\addlinespace[1pt]\cline{2-22}\addlinespace[1pt]

& \multicolumn{1}{c|}{\multirow{2}{*}{UEPS}} & Short & 0.083 & 0.085 & 0.091 & 0.086 & 0.093 & 0.087 & 0.087 & 0.086 & 0.087 & 0.085 & \secondbest{0.083} & 0.089 & 0.088 & 0.095 & 0.105 & 0.093 & \best{0.080} & 0.470 & 0.088 \\
& & Long & \secondbest{0.132} & 0.132 & 0.135 & 0.136 & 0.138 & 0.142 & 0.139 & 0.158 & 0.151 & 0.133 & 0.141 & 0.137 & \best{0.131} & 0.144 & 0.145 & 0.151 & 0.135 & 0.135 & 0.151 \\
\addlinespace[1pt]\cline{2-22}\addlinespace[1pt]

& \multicolumn{1}{c|}{\multirow{2}{*}{UEBB}} & Short & 0.107 & 0.107 & 0.107 & 0.109 & 0.113 & 0.108 & 0.117 & 0.114 & 0.110 & 0.109 & 0.110 & 0.109 & 0.107 & 0.112 & 0.112 & \secondbest{0.105} & \best{0.103} & 0.110 & 0.110 \\
& & Long & \secondbest{0.141} & \best{0.141} & 0.144 & 0.147 & 0.147 & 0.150 & 0.163 & 0.199 & 0.159 & 0.148 & 0.154 & 0.148 & 0.145 & 0.150 & 0.149 & 0.168 & 0.142 & 0.148 & 0.155 \\

\bottomrule
\end{tabular}%
}
\vspace{2pt}

\footnotesize MT1V/MTMV-only nMAE results including multi-turbine-only models. Short averages horizons 12 and 24; Long averages horizons 72 and 144. Red bold entries denote the best results, and blue italic underlined entries denote the second-best results in each dataset--horizon-group row; ``--'' indicates unavailable results.
\end{table*}

For cross-dataset comparison, we further use a GIFT-style normalized evaluation. Each model score is divided by the
SeasonalNaive score on the same structure--dataset--horizon unit,
then aggregated by a geometric mean within each structure and by
a structure-balanced summary across applicable structures. 

\begin{figure*}[!t]
\centering
\includegraphics[width=\textwidth]{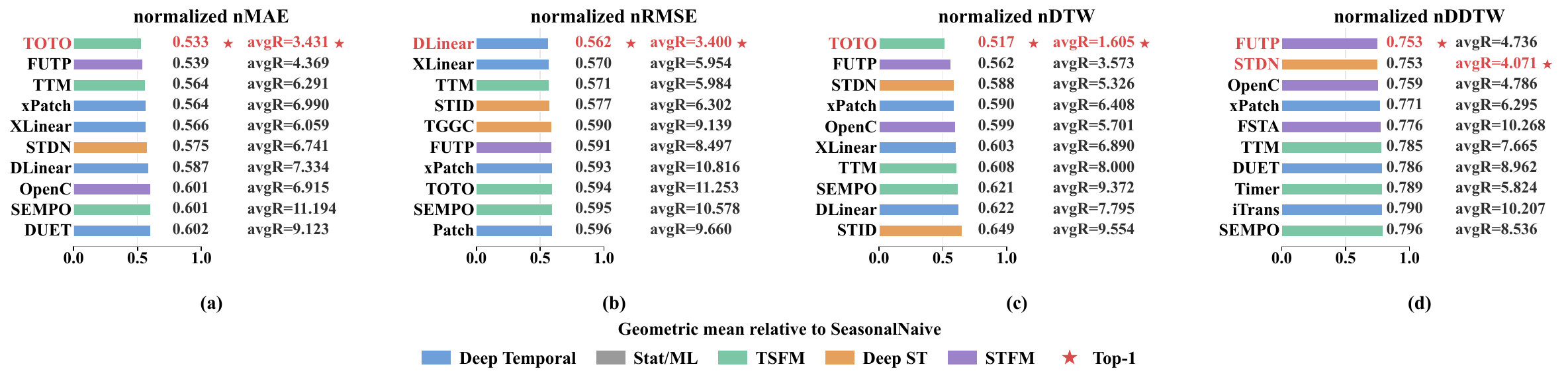}
\caption{Normalized leaderboard under SeasonalNaive-based scaling.
Bars show structure-balanced geometric means relative to SeasonalNaive,
and avgR labels report scope-aware average ranks.}
\label{fig:ch5_gift_normalized_metric_rank_bars}
\end{figure*}

Figure~\ref{fig:ch5_gift_normalized_metric_rank_bars} summarizes
two representative normalized leaderboards, using nMAE for
numerical accuracy and nDTW for forecast-curve fidelity. Under
nMAE, foundation models such as TOTO,
FactoST-UTP, and TTM achieve the best normalized scores. Under
nDTW, TOTO and FactoST-UTP remain highly competitive, while
larger and more structurally complex spatio-temporal models such
as STDN and OpenCity also move into the leading group. Meanwhile,
several linear/MLP-style temporal models that perform strongly
under point-wise errors, such as DLinear and TTM, show some degree
of rank degradation under forecast-curve fidelity evaluation.

The remaining normalized metrics further support this observation.
For numerical error metrics, nRMSE tends to favor linear/MLP-style
temporal models such as DLinear, XLinear, and TTM, possibly
because these models produce smoother forecasts and suppress large
prediction deviations, as nRMSE is more sensitive to extreme errors.
Overall, the best model under numerical accuracy is not necessarily
the best model under forecast-curve fidelity. WPBench therefore
reports these two perspectives separately rather than merging them
into a single score.

\subsection{Structure-aware Diagnostic Analysis}
To further understand how dataset characteristics affect forecasting behavior,
we analyze model responses using the characterization metrics defined in
Section~\ref{sec:dataset_characterization_metrics}, covering temporal patterns,
variable dependencies, and spatial dependencies. For each metric, datasets are ranked by the corresponding metric value, and LowQ and HighQ groups are formed from the bottom and top 20\%, respectively. To reduce the influence of individual horizons and outlier datasets, we report shifted-geometric-mean nMAE aggregated
over the selected dataset--horizon units.

\begin{figure*}[!t]
\centering
\includegraphics[width=0.9\textwidth]{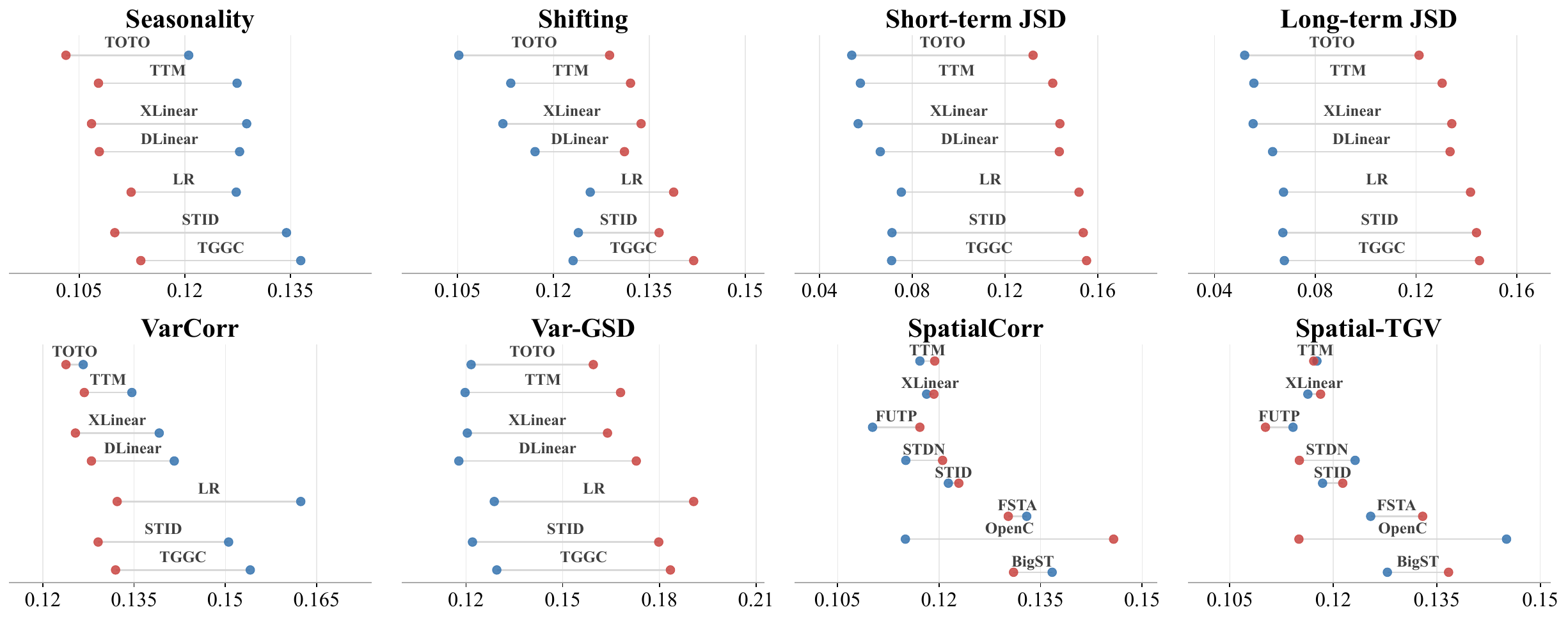}
\caption{Extreme dataset-characteristic w.r.t representative models.
Blue and red markers denote shifted-geometric-mean nMAE on LowQ and HighQ
datasets, respectively. The distance between markers indicates the model's
response difference to the corresponding dataset characteristic. Lower nMAE
is better. LowQ and HighQ are formed by the bottom and top 20\% datasets
ranked by each characterization metric.}
\label{fig:ch5_extreme_selected_model_response}
\end{figure*}

Figure~\ref{fig:ch5_extreme_selected_model_response} first shows that temporal characteristics affect most model families in a similar direction. Strong seasonality generally corresponds to lower forecasting error, suggesting that periodic wind-power patterns provide exploitable regularity. In contrast, metrics related to distributional shift and temporal divergence tend to increase error, indicating that non-stationary temporal patterns remain a major difficulty for wind-power forecasting. This contrast suggests that temporal regularity and temporal drift should be treated separately: periodicity is useful, while changing operating regimes or distributional divergence can weaken model robustness.

We next examine variable-dependency patterns. High VarCorr can reduce forecasting error for several representative models, showing that stable target--covariate relationships provide useful predictive information. However, when variable relationships become more dynamic, as reflected by Var-GSD, the benefit becomes less stable. This indicates that multivariate information is helpful only when the dependency pattern is sufficiently consistent; otherwise, models may capture transient or noisy correlations.

\begin{figure}[t]
\centering
\includegraphics[width=\columnwidth]{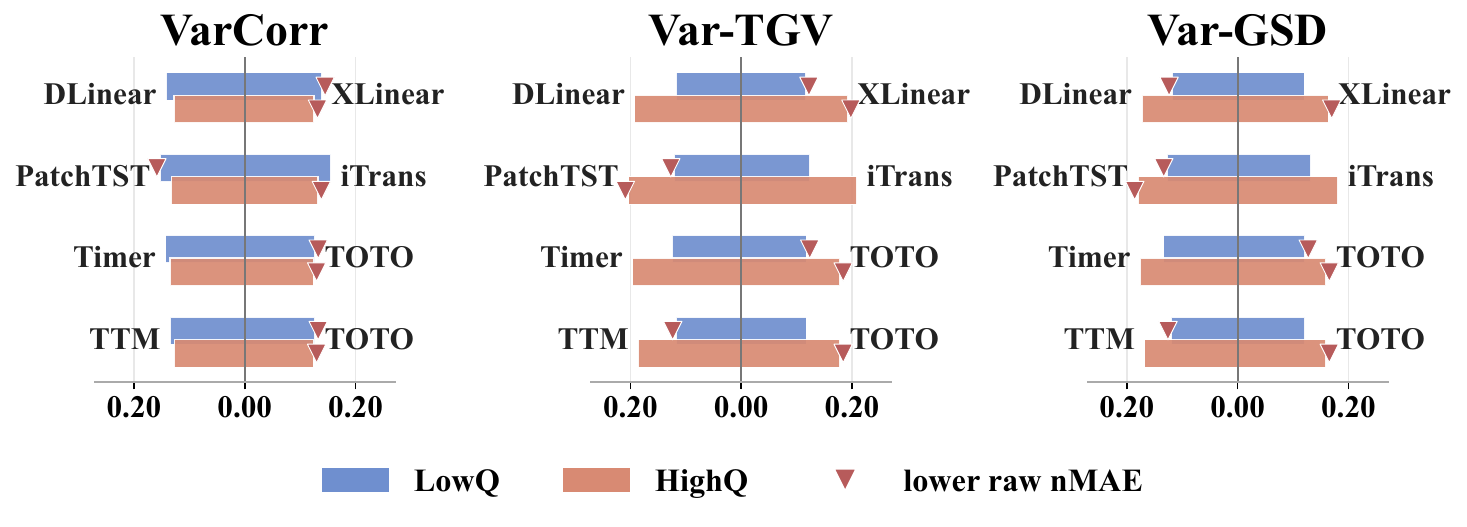}
\caption{Paired CI/CD model comparison w.r.t. variable-dependency conditions. Each panel corresponds to one variable-dependency descriptor, and each row compares a matched channel-independent (CI) and channel-dependent (CD) model pair. Bars extending left and right report the nMAE of the CI and CD models, respectively, under LowQ and HighQ groups; red triangles mark the lower-error side.}
\label{fig:ch5_variable_ci_cd_response}
\end{figure}

Figure~\ref{fig:ch5_variable_ci_cd_response} further compares representative channel-independent and channel-dependent model pairs. Under stronger or more dynamic variable relationships, some channel-dependent models outperform their structurally similar channel-independent counterparts, indicating that explicit channel-dependency modeling can help exploit information from relationships between the target and covariates. However, this advantage is not universal: although iTransformer performs competitively under VarCorr, it weakens under Var-TGV and Var-GSD and can be outperformed by PatchTST. This suggests that variable dependency is useful, but current channel-dependent designs still need stronger dynamic relation modeling to exploit it reliably.

Finally, we analyze spatial dependency characteristics. Unlike temporal
periodicity, spatial characteristics do not lead to a simple monotonic
conclusion. SpatialCorr measures static cross-turbine correlation, whereas
Spatial-TGV reflects how spatial relations change over time. As shown in
Figure~\ref{fig:ch5_extreme_selected_model_response}, higher spatial dependency
does not consistently correspond to lower forecasting error. Instead, it may
also reveal model sensitivity to unstable or mismatched spatial assumptions.

\begin{figure}[t]
\centering
\includegraphics[width=\columnwidth]{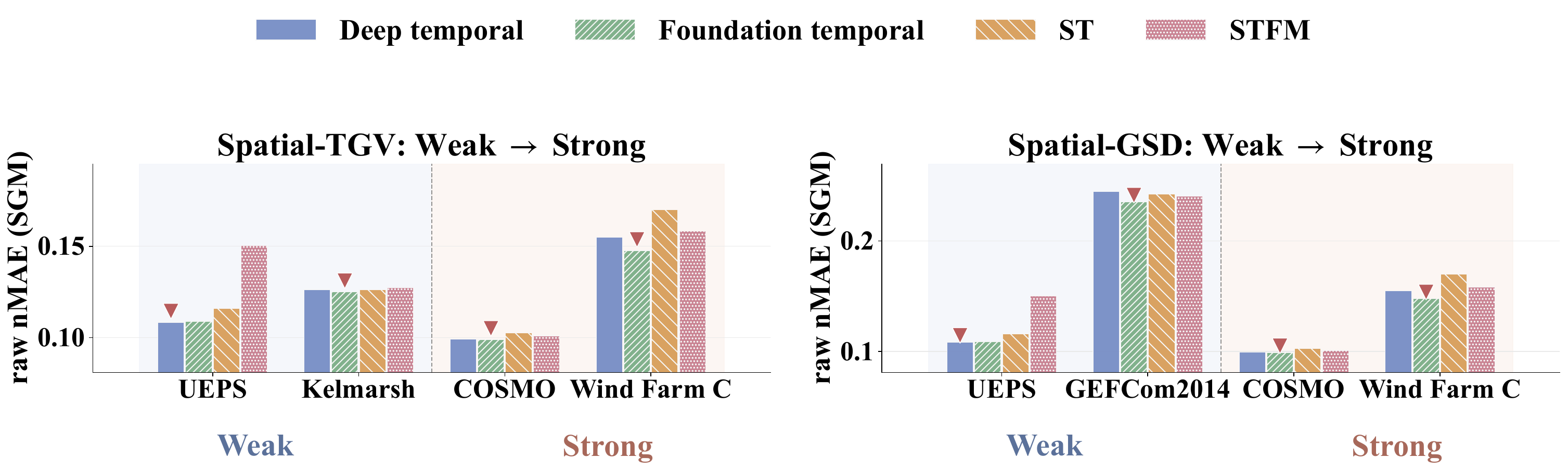}
\caption{Dataset-level spatial w.r.t model group. Bars report shifted-geometric-mean nMAE for temporal, foundation temporal, deep spatio-temporal, and spatio-temporal foundation model groups on weak and strong spatial-dynamic datasets. Red triangles mark the best group on each dataset.}
\label{fig:ch5_spatial_dataset_group_response}
\end{figure}

Figure~\ref{fig:ch5_extreme_selected_model_response} and Figure~\ref{fig:ch5_spatial_dataset_group_response} show that spatially aware models are often more sensitive to changes in spatial dependency patterns. Models such as OpenCity, FactoST-STA, STDN, and BigST react more strongly to spatial descriptors than purely temporal references, but their responses are not uniform. One possible explanation is that spatial relations in wind farms are less stable than the road-network structures commonly assumed in traffic forecasting. These relations can vary with wind conditions, terrain, wake effects, and turbine operating states. As a result, spatial modeling can help when its assumptions match the data, but it may also amplify errors when the learned spatial prior is unstable or mismatched.

Overall, these diagnostics show that WPBench contains diverse dataset
characteristics rather than a single homogeneous forecasting setting. Temporal periodicity is generally beneficial, variable dependency is useful but requires robust dynamic modeling, and spatial dependency introduces additional sensitivity rather than guaranteed gains. This explains why the overall leaderboard does not simply favor spatially complex models: models that ignore unstable spatial relations can sometimes be more robust, while spatial models require better adaptation to wind-specific dependency dynamics.

\subsection{Numerical Accuracy vs. Forecast-Curve Fidelity}
\label{sec:numerical_accuracy_vs_forecast_curve_fidelity}

\begin{figure}[t]
\centering
\includegraphics[width=\columnwidth]{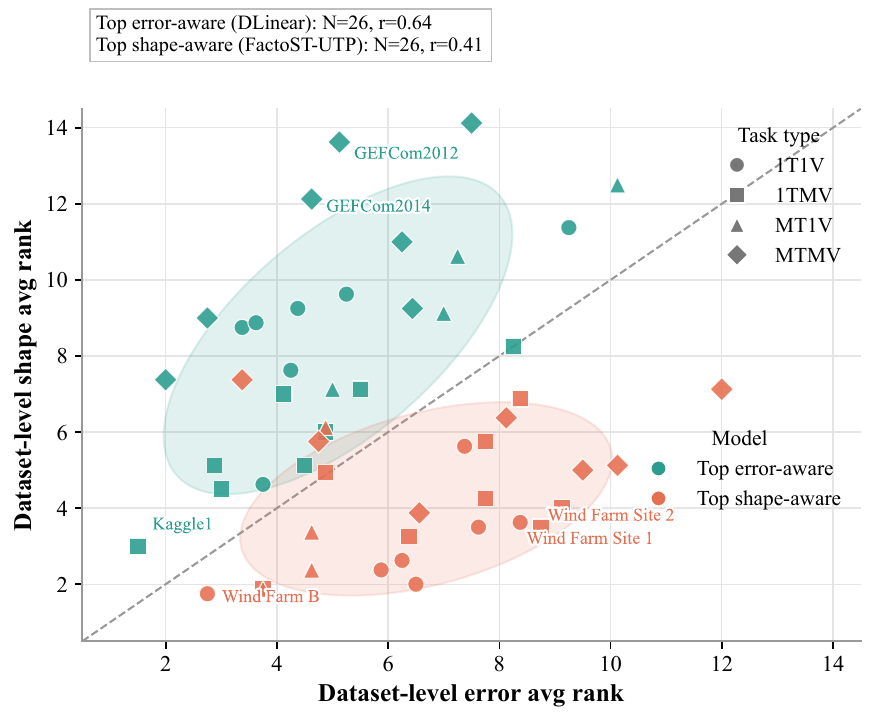}
\caption{Dataset-level comparison between Error Rank and Shape Rank for
representative models. Error Rank averages ranks over nMAE and nRMSE, and
Shape Rank averages ranks over nDTW and nDDTW. Shaded regions summarize the
rank distributions of the top-ranked models under numerical accuracy and
forecast-curve fidelity.}
\label{fig:error_shape_profiles}
\end{figure}

We further examine whether numerical accuracy and forecast-curve fidelity lead
to consistent model conclusions. Figure~\ref{fig:error_shape_profiles} compares
the dataset-level Error Rank and Shape Rank of the top-ranked models under
these two evaluation views. The two representatives show clearly different rank
distributions: the top-ranked model under numerical accuracy usually maintains
stronger numerical accuracy, but its Shape Rank can degrade; the top-ranked
model under forecast-curve fidelity better preserves curve similarity, but its
Error Rank varies more substantially. This suggests that current models often
favor one evaluation objective and struggle to simultaneously achieve both
numerical accuracy and forecast-curve fidelity.

\begin{figure}[!t]
\centering
\includegraphics[width=1\columnwidth]{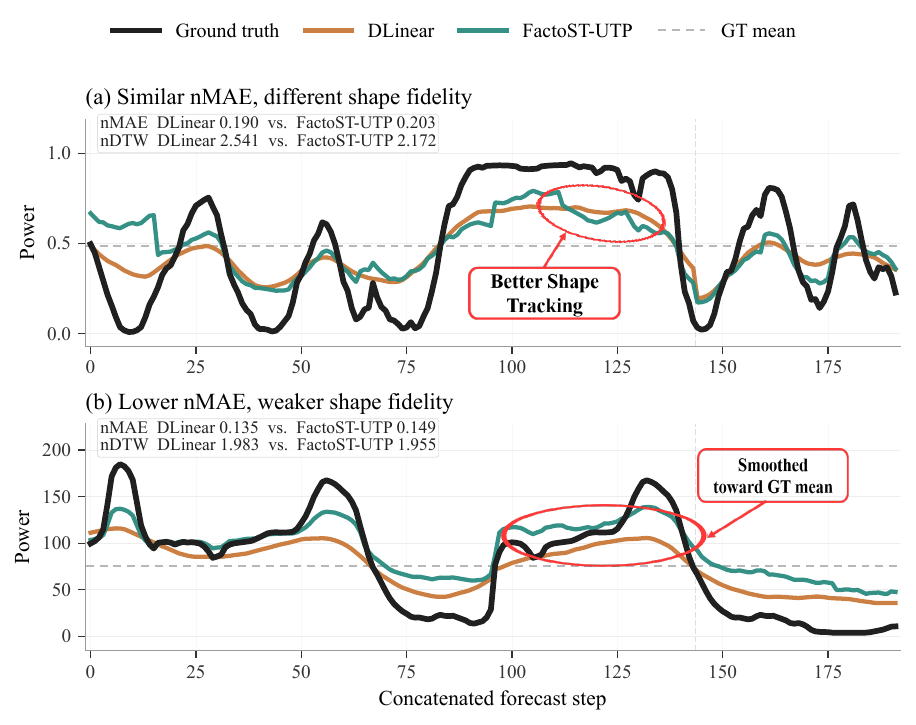}
\caption{Forecast-curve case study on GEFCom2012.}
\label{fig:error_shape_case_study}
\end{figure}

Figure~\ref{fig:error_shape_case_study} further shows a representative forecast
segment on GEFCom2012. DLinear is closer to the average level of the true power
curve and therefore obtains lower nMAE, but its forecasts are overly smooth and
tend to miss local ramps, short-term fluctuations, and peak and trough
variations. FactoST-UTP better follows the temporal shape of the true sequence
in some segments and thus obtains lower nDTW; however, it can still exhibit
temporal lag, excessive fluctuation, or amplitude offset, which increases
point-wise error.

These results show that lower numerical error does not necessarily imply better
forecast-curve fidelity. WPBench therefore reports Error Rank and Shape Rank as
complementary views rather than merging them into a single overall ranking.

\subsection{Foundation Model Adaptation}

\begin{figure}[!t]
\centering
\includegraphics[width=1\columnwidth]{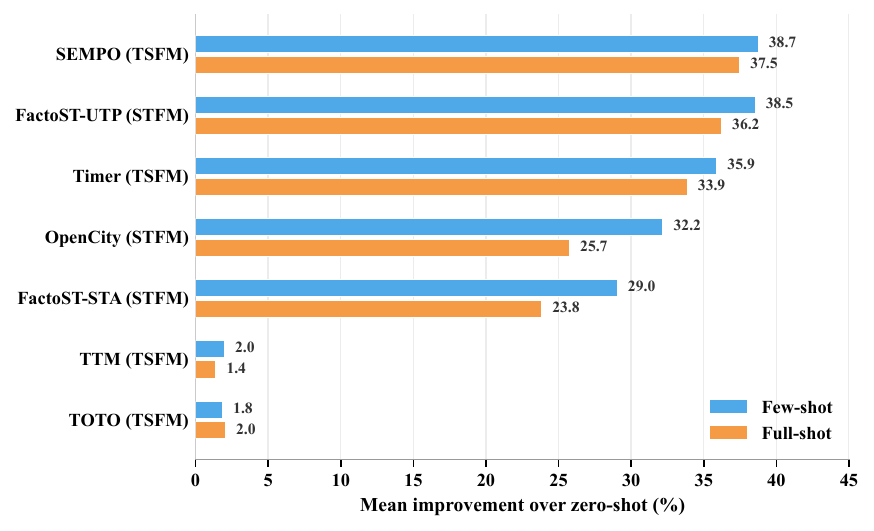}
\caption{Model-level foundation adaptation summary. Bars show the relative nMAE reduction of few-shot and full-shot adaptation over zero-shot performance, averaged over each model's applicable wind data structures.}
\label{fig:foundation_adaptation_summary}
\end{figure}

We compare temporal and spatio-temporal foundation models under zero-shot,
few-shot (10\% uniformly sampled training data), and full-shot settings.
Figure~\ref{fig:foundation_adaptation_summary} reports the relative nMAE
reduction of each model after adaptation, using its zero-shot performance as the reference.

Figure~\ref{fig:foundation_adaptation_summary} shows that adaptation gains are largest for models with ample zero-shot calibration room (SEMPO, FactoST-UTP, Timer), while TTM and TOTO improve little, likely due to strong zero-shot baselines. Notably, many models achieve few-shot gains comparable to or slightly above full-shot, suggesting that a small representative subset can capture key wind-farm patterns, whereas full-shot introduces more low-wind, volatile, curtailed, or abnormal segments. Adaptation thus depends on sample representativeness, regularization, and validation, not just data volume.

Beyond data volume, FactoST-UTP outperforms STA, suggesting that stronger cross-turbine aggregation does not necessarily improve adaptation. Given the complex and non-stationary spatio-temporal dependencies in wind farms, joint channel modeling may overfit unstable spatial relations, whereas channel-independent adaptation can be more robust.

\subsection{Model Efficiency}
\begin{figure}[!t]
\centering
\includegraphics[width=1\columnwidth]{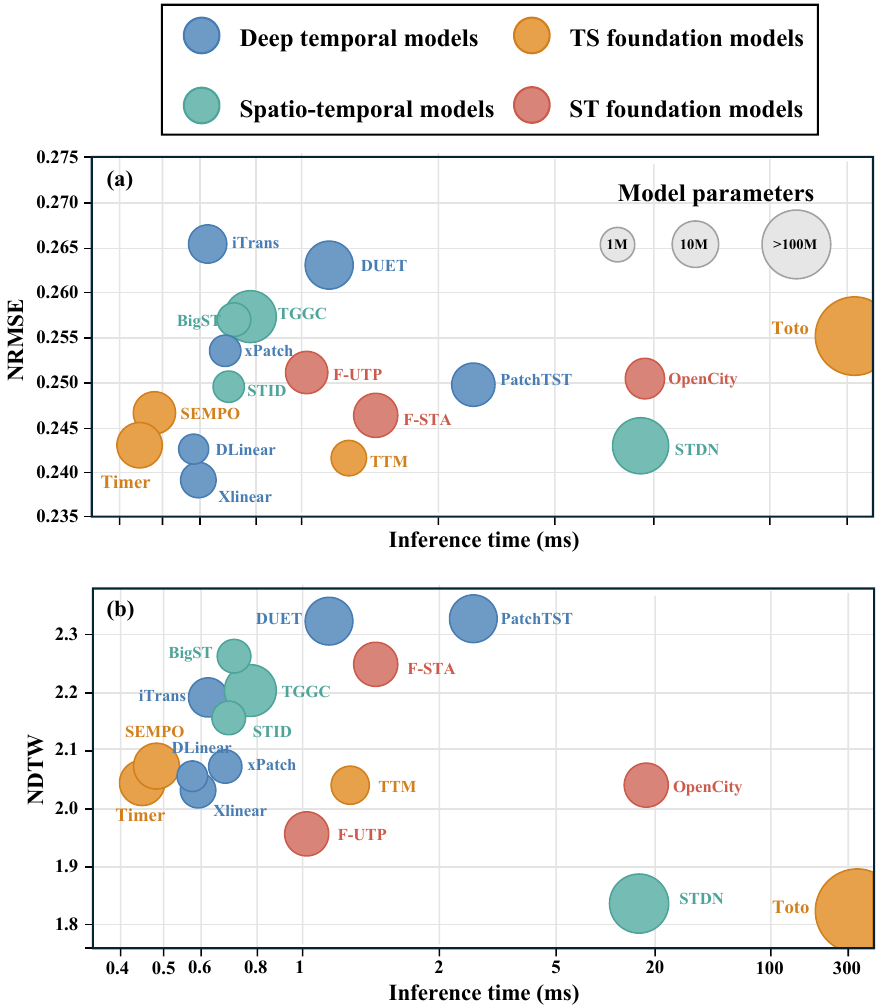}
\caption{Efficiency--accuracy on GEFCom2012 with horizon 12. Bubble size denotes the number of model parameters. Panel (a) compares inference time with nRMSE, while Panel (b) compares inference time with nDTW.}
\label{fig:ch5_efficiency_dual_metric}
\end{figure}
Figure~\ref{fig:ch5_efficiency_dual_metric} compares the efficiency--accuracy trade-off from the perspectives of numerical accuracy and forecast-curve fidelity. In Figure~\ref{fig:ch5_efficiency_dual_metric}(a), lightweight linear or MLP-style models such as DLinear, XLinear, STID, and TTM achieve low inference cost with competitive nRMSE, suggesting that simple architectures can already provide stable numerical forecasts. This is consistent with Sec.~\ref{sec:numerical_accuracy_vs_forecast_curve_fidelity}, where DLinear tends to produce smoother predictions and thus more stable numerical errors. In Figure~\ref{fig:ch5_efficiency_dual_metric}(b), however, larger or more complex models such as STDN and TOTO show stronger forecast-curve fidelity under nDTW, indicating that models with larger capacity, especially spatio-temporal and time-series foundation models, are more capable of preserving temporal dynamics and curve shape. Notably, TOTO achieves strong shape-aware performance but incurs much higher inference latency, which may be partly related to its probabilistic forecasting interface that produces quantile outputs or sample-based outputs before deriving point forecasts for evaluation. Under rolling evaluation, this additional inference cost becomes more visible because forecasts must be repeatedly generated for many test windows. Such latency can be less suitable for scenarios with strict real-time requirements, such as frequent rolling updates, online dispatch, or large-scale turbine monitoring. In contrast, faster models may be more suitable for deployment when a small sacrifice in curve fidelity leads to substantial savings in response time. Overall, the figure highlights a trade-off between efficient numerical prediction and stronger forecast-curve modeling at a higher computational cost.


\section{Conclusions}
This paper presents WPBench, a comprehensive bench- mark for wind power forecasting across diverse wind data structures. WPBench evaluates a broad range of forecasting models, including statistical baselines, deep temporal mod- els, spatio-temporal models, time-series foundation models, and spatio-temporal foundation models, under unified pro- tocols. Beyond conventional point-wise accuracy, WPBench incorporates shape-aware metrics for forecast-curve fidelity, evaluates computational efficiency, and provides structure- aware diagnostics from temporal, variable-dependency, and spatial-dependency perspectives. Through diagnostic analy- sis, we show that model performance is strongly affected by wind data organization, variable dependency, and spatial structure. Overall, WPBench provides a reproducible eval- uation framework for understanding the strengths and lim- itations of existing forecasting models and offers guidance for developing more robust wind-specific forecasting methods. 


\bibliographystyle{IEEEtran}
\bibliography{references}

\end{document}